%% file: main.tex
\documentclass[]{article} %% [draft]
\usepackage{arxiv}

\usepackage[T1]{fontenc}
\usepackage[utf8]{inputenc}
\usepackage{nicefrac}
\usepackage[font=small,labelfont=bf]{caption}
\usepackage{subcaption}
\usepackage{parskip}
\usepackage{float}
\usepackage{graphicx}
\usepackage{amsfonts}
\usepackage{amsmath}
\usepackage{amssymb}
\usepackage{mathtools}
\usepackage{upgreek}
\usepackage{siunitx}
\usepackage[final]{showkeys}
\usepackage{hyperref}
\usepackage{url}
\usepackage[authoryear,longnamesfirst]{natbib}
\usepackage{authblk}
\usepackage{booktabs,makecell,multirow,array,colortbl,dcolumn,stfloats}

\graphicspath{{./}}
\DeclareSIUnit[quantity-product = {}]
\ppm{\text{ppm}}
\DeclarePairedDelimiterX{\abs}[1]{\lvert}{\rvert}{#1}
\DeclarePairedDelimiter{\ceil}{\lceil}{\rceil}
\DeclarePairedDelimiter{\flr}{\lfloor}{\rfloor}
\newcommand{\tr}{{\mkern-5mu \top}}
\newcommand{\inR}[1]{\in\mathbb{R}^{#1}}
\newcommand{\inC}[1]{\in\mathbb{C}^{#1}}
\newcommand{\diff}{\mbox{\,d}}
\newcommand{\dotp}[3]{\ensuremath{\left\langle {#1}, {#2} \right\rangle_{#3}}}
\newcommand{\ie}{\textit{i.e.} }
\newcommand{\dquotes}[1]{``{#1}''}
\newcommand{\eref}[1]{\eqref{#1}} 
\newcommand{\fref}[1]{Fig.~\ref{#1}}
\newcommand{\tref}[1]{Tab.~\ref{#1}}
\newcommand{\sref}[1]{Sec.~\ref{#1}}
\DeclareMathOperator{\diag}{Diag}
\DeclareMathOperator*{\argmin}{arg\,min}% thin space, limits underneath in displays
\def\norm#1{\left\|#1\right\|}

\graphicspath{{./}}
\title{End-to-end data-driven prediction of urban airflow and pollutant dispersion}

\date{}

\newbox{\orcid}\sbox{\orcid}{} %{\includegraphics[scale=0.06]{orcid}} 
\author[1]{%
	\href{https://orcid.org/0000-0001-5759-0708}{\hspace{1mm}Nishant Kumar\thanks{\texttt{nishant.kumar@ec-nantes.fr}}}% 
}
\author[1]{%
    \href{https://orcid.org/0000-0003-0299-6991}{\hspace{1mm}Franck Kerherv\'{e}}%
}
\author[1]{%
    \href{https://orcid.org/0000-0001-8305-5493}{\hspace{1mm}Lionel Agostini}%
}
\author[1]{%
    \href{https://orcid.org/0000-0002-8085-6102}{\hspace{1mm}Laurent Cordier}%
}
\affil[1]{Institut Pprime, CNRS, Universit\'e de Poitiers, ISAE-ENSMA, Curiosity Team, B\^at. H2 - Bd. Marie \& Pierre Curie TSA 51124, 86073 Poitiers Cedex 9, France}

\renewcommand{\headeright}{Preprint}
\renewcommand{\undertitle}{Preprint}
\renewcommand{\shorttitle}{End-to-end data-driven prediction of pollutant dispersion}

\hypersetup{
pdftitle={End-to-end data-driven prediction of urban airflow and pollutant dispersion},
pdfsubject={physics.flu-dyn},
pdfauthor={Nishant Kumar},
pdfkeywords={urban airflow, pollutant dispersion, reduced-order modeling },
}

\begin{document}
\maketitle

%% Abstract
\begin{abstract}
Climate change and the rapid growth of urban populations are intensifying environmental stresses within cities, making the behavior of urban atmospheric flows a critical factor in public health, energy use, and overall livability. This study targets to develop fast and accurate models of urban pollutant dispersion to support decision-makers, enabling them to implement mitigation measures in a timely and cost-effective manner.
To reach this goal, an end-to-end data-driven approach is proposed to model and predict the airflow and pollutant dispersion in a street canyon in skimming flow regime.
A series of time-resolved snapshots obtained from large eddy simulation (LES) serves as the database.
The proposed framework is based on four fundamental steps.
Firstly, a reduced basis is obtained by spectral proper orthogonal decomposition (SPOD) of the database.
The projection of the time series snapshot data onto the SPOD modes (time-domain approach) provides the temporal coefficients of the dynamics.
Secondly, a nonlinear compression of the temporal coefficients is performed by autoencoder to reduce further the dimensionality of the problem.
Thirdly, a reduced-order model (ROM) is learned in the latent space using Long Short-Term Memory (LSTM) netowrks. 
Finally, the pollutant dispersion is estimated from the predicted velocity field through convolutional neural network that maps both fields.
The results demonstrate the efficacy of the model in predicting the instantaneous as well as statistically stationary fields over long time horizon.
\end{abstract}

%% Keywords
\keywords{
Reduced-order modeling \and
Data-driven approach \and
Neural networks \and
Deep Learning \and
Urban airflow \and
Pollutant dispersion 
}

%% Uncomment to enable line numbers
% \linenumbers

%% Main text
\input{section-intro.tex}
\input{section-cfd-simulation.tex}
\input{section-method.tex}
\input{section-results.tex}
\input{section-conclusion.tex}

\section*{CRediT authorship contribution statement}
\textbf{Nishant Kumar}: Writing – review and editing, Writing – original draft, Methodology, Investigation, Conceptualization. 
\textbf{Franck Kerherv\'{e}}: Writing – review and editing, Supervision, Funding acquisition. 
\textbf{Lionel Agostini}: Writing – review and editing, Supervision, Methodology, Funding acquisition, Data curation.
\textbf{Laurent Cordier}: Writing – review and editing, Supervision, Methodology, Funding acquisition, Data curation.

% \section*{Declaration of competing interest}
% The authors declare the following financial interests/personal relationships which may be considered as potential competing interests:
% Laurent Cordier reports financial support was provided by French National Research Agency.

\section*{Data availability}
The data that support the findings of this study are available from the corresponding author upon reasonable request.

\section*{Acknowledgements}
This material is based upon work supported by the French National Research Agency through the research grant ANR-22-CE22-0008 for the \dquotes{Modeling Urban Flows using Data-Driven methods} -- MUFDD project.

\bibliographystyle{cas-model2-names}
\bibliography{references}

\end{document}

%% file: section-intro.tex
% ---------------------------------------------------------------------------
\section{Introduction}
% ---------------------------------------------------------------------------
\label{sec:intro}

The air quality in cities is adversely affected by air pollution, posing a direct risk to public health.
The dispersion of air pollutants is primarily driven by the wind interacting with the urban morphology \citep{liReviewPollutantDispersion2021}, which makes the placement of buildings a key point of concern in urban development \citep{linReviewUrbanPlanning2024}.
Long rows of buildings are often encountered in cities, forming regions known as \emph{urban street canyons}, which are similar in principle to the canonical cavity configuration.
Based on the {aspect ratio} of the canyon, \ie the ratio between building height and street width, different airflow regimes are observed.
High-density urban areas are commonly characterized by an aspect ratio larger than 0.65 \citep{okeBoundaryLayerClimates2002}.
The flow in these canyons is categorized as \emph{skimming flow}, where the roof-level wind skims over the top surface of the buildings and drives a vortex in the cavity, which in turn influences the removal of pollutants at street-level.

While airflow prediction in urban areas is critical for satisfying air quality requirements, the complexity of urban geometries, the variability of meteorological conditions and the multiscale nature of turbulent flows make it challenging \citep{hubayTurbulentFlowStreet2025}.
Computational fluid dynamics (CFD) tools, especially large eddy simulation (LES), have been used extensively to simulate urban airflow with high fidelity \citep{cuiLargeeddySimulationTurbulent2004}.
However, CFD models commonly incur very high computational costs and are often inapplicable for real-time forecasting or extensive parametric studies \citep{torresExperimentalNumericalDataDriven2021}.
The objective of this paper is then to develop a numerical pipeline that can efficiently predict pollutant dispersion for turbulent flow representative of urban canopy flows. 

Model order reduction approaches offer the potential to reduce the degrees of freedom of the system while preserving key physical features \citep{masoumi-verkiReviewAdvancesEfficient2022}, thereby offering a low-cost and tractable predictive tool for air quality surveys.
Data-driven strategies have increasingly gained traction as tools for non-intrusive reduced-order modeling (ROM) in fluid dynamics problems related to urban flows \citep{perretCombiningWindTunnelField2016,xiaoReducedOrderModel2019}.
The fundamental challenge of any ROM is the trade-off between finding a latent space that accurately represents the high-dimensional physics and keeping the dimensionality sufficiently low for computational efficiency.
Proper orthogonal decomposition (POD) has been used extensively to derive reduced basis functions that are optimal in terms of energy content \citep{waltonReducedOrderModelling2013}.
However, standard POD modes often mix temporal scales, which can be limiting for urban flows characterized by distinct frequency mechanisms.
To address this, the spectral POD (SPOD) approach proposed by \cite{towneSpectralProperOrthogonal2018} is considered in this work.
For statistically stationary data, SPOD combines the optimality of POD with the ability to isolate coherent structures at specific frequencies \citep{schmidtGuideSpectralProper2020}.
This capability is particularly advantageous for street canyon flows, which exhibit distinct multiscale dynamics, such as shear layer instabilities and vortex shedding \citep{zhangSpectralProperOrthogonal2022}.

While SPOD effectively isolates spectrally coherent structures, accurate reconstruction of turbulent flows often requires retaining a large number of modes, resulting in a high-dimensional modal space that can hinder efficient time forecasting \citep{schmidtGuideSpectralProper2020}.
To surmount this limitation, we employ a dense autoencoder (AE) to discover a compact, nonlinear latent representation of the SPOD coefficients.
Borrowing from the nonlinear POD framework proposed by \cite{ahmedNonlinearProperOrthogonal2021}, the AE serves as a secondary, high-efficiency compressor, capturing the manifold of the dynamics more effectively than linear truncation alone \citep{agostiniExplorationPredictionFluid2020}.

For the temporal evolution of this compressed latent space, recurrent neural network (RNN) architectures have been proven to be efficient in handling time-series data
\citep{caoAutomaticSelectionBest2025}.
Unlike traditional autoregressive (AR) models which assume linear relationships, RNNs are adapted to the nonlinear dynamics inherent in fluid flows.
Specifically, the long short-term memory (LSTM) network \citep{hochreiterLongShortTermMemory1997} is utilized here, as it is immune to the vanishing gradient problem affecting classic RNNs.
LSTM models have been successfully employed to propagate latent space variables in POD \citep{ahmedNonlinearProperOrthogonal2021} and SPOD  frameworks \citep{larioNeuralnetworkLearningSPOD2022}.
In the proposed modular framework, the LSTM forecasts the dynamics of the latent space obtained by encoding the SPOD coefficients using AE, while the decoder maps these predictions back to the SPOD representation. 

Finally, to predict pollutant dispersion, a mapping between the flow field and the passive scalar field is required.
We posit that for continuous sources in a statistically stationary regime, the instantaneous concentration field is strongly correlated with the driving velocity structures.
Consequently, we employ a convolutional neural network (CNN) \citep{lecunGradientbasedLearningApplied1998} to learn this complex mapping.
CNNs have demonstrated their ability to reconstruct flow features from sparse data \citep{guastoniConvolutionalnetworkModelsPredict2021,kimUnsupervisedDeepLearning2021} and offer a robust method for velocity-to-scalar mapping \citep{nakamuraRobustTrainingApproach2022}.

The primary objective of this work is to develop a modular, end-to-end data-driven ROM framework that addresses the specific challenges of predicting pollutant in the skimming regime of urban canyons.
By integrating SPOD for spectral isolation, AE for nonlinear compression, LSTM for temporal forecasting, and CNN for scalar transport, we propose a robust pipeline capable of predicting both instantaneous fields and long-term statistics.
The paper is organized as follows: \sref{sec:numsim} presents the numerical dataset obtained from the street canyon LES, \sref{sec:method} describes the methodology for each component of the framework, \sref{sec:result} illustrates the performance of the ROM while reproducing high-fidelity solutions and forecasting dynamics, and conclusions are highlighted in \sref{sec:conc}.

%% file: section-cfd-simulation.tex
% ---------------------------------------------------------------------------
\section{Numerical dataset}
% ---------------------------------------------------------------------------
\label{sec:numsim}

\begin{figure}
    \centering
    \begin{subfigure}{0.35\textwidth}
        \centering
        \includegraphics[width=\textwidth]{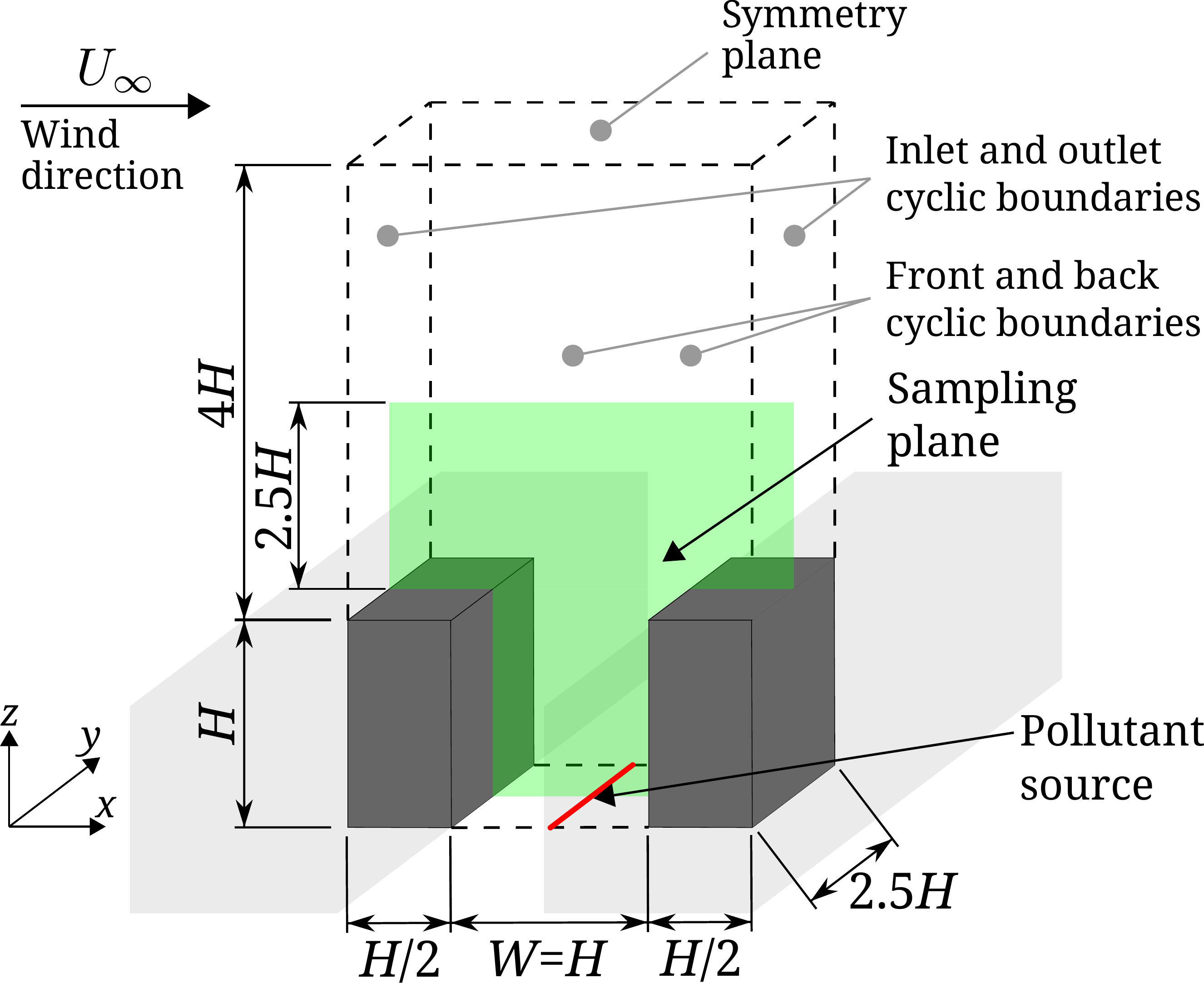}
        \caption{}
        \label{fig:numsim:1a}
    \end{subfigure}
    \begin{subfigure}{0.3\textwidth}
        \centering
        \includegraphics[width=\textwidth]{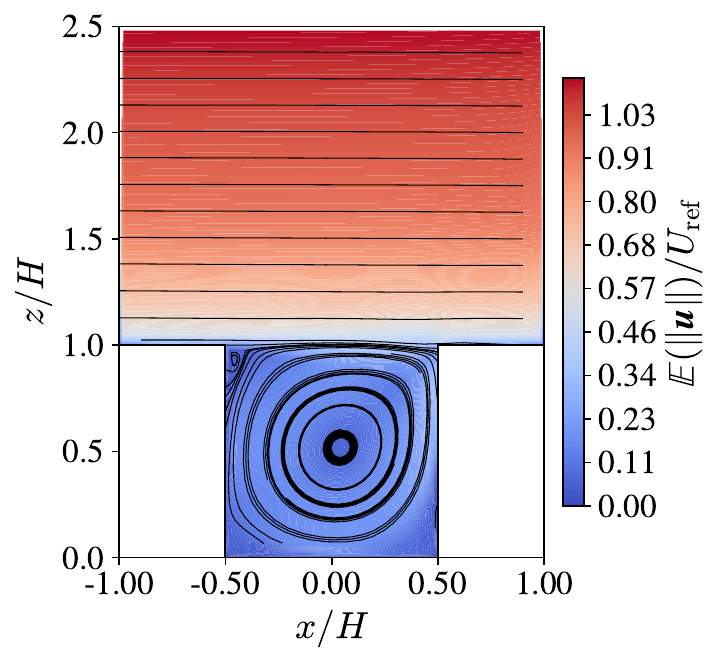}
        \caption{}
        \label{fig:numsim:1b}
    \end{subfigure}
    \begin{subfigure}{0.3\textwidth}
        \centering
        \includegraphics[width=\textwidth]{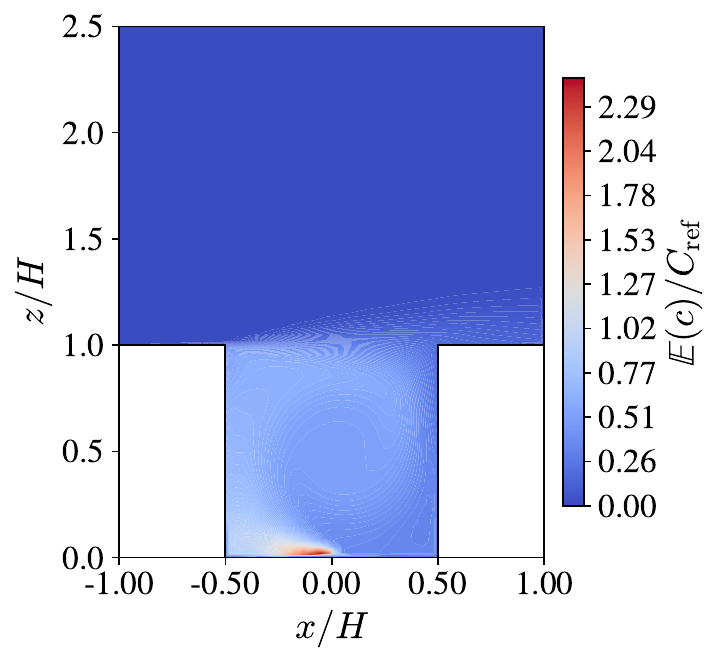}
        \caption{}
        \label{fig:numsim:1c}
    \end{subfigure}
    \caption{
    (a) Overview of the simulated street canyon geometry (not to scale).
    The bounds of the computational domain is indicated by dashed line along with the boundary conditions.
    The pollutant source is indicated by the red line which spans the center of the canyon base.
    The vertical sampling plane located mid-span of the canyon is indicated in green.
    (b) Mean velocity magnitude with streamlines on the $x$-$z$ plane.
    (c) Mean concentration of the pollutant emitted from source.
    }
    \label{fig:numsim:1}
\end{figure}

A large eddy simulation (LES) of a 3D street canyon flow was performed by \citet{combettePODGalerkinReducedorder2026} using OpenFOAM (v2306). 
The simulation considered an \emph{ideal} configuration as shown in \fref{fig:numsim:1a}.
Two rows of rectangular blocks of dimensions $H \times 0.5H \times 2.5H$ (height $\times$ width $\times$ length) with $H=\qty{0.1}{\meter}$ are placed such that the gap between the blocks forms a canyon of width $W$ and an aspect ratio of $H/W = 1$.
Additionally, a continuous line source was added at the center of the base of the canyon to represent pollutant emissions.

Recycling-rescaling boundary conditions were applied on the inlet and outlet in the streamwise direction ($x$), and periodic boundary conditions on the two sides in the span-wise direction ($y$).
This allows upstream conditions representative of urban flows to be obtained.
Moreover, the wind flow direction was considered perpendicular to the span of the canyon.
The resulting symmetry means that the flow can be considered as quasi-2D.

The values of streamwise and vertical components of velocity $\mathbf{u}(\mathbf{x}, t) = [u(\mathbf{x}, t), w(\mathbf{x}, t)]^\tr$ and scalar concentration $c(\mathbf{x}, t)$ were sampled in the domain
\begin{equation}
    \Omega = \left\{ \mathbf{x} = (x,y,z)^\tr \vert -H \leq x \leq H, \, y=0, \, 0 \leq z \leq 2.5H \right\},
    \label{eq:numsim:1}
\end{equation}
on a uniform grid of resolution $H/30$ containing $N_x = 7200$ points.
The sampling plane is highlighted in \fref{fig:numsim:1a}.
The time-averaged streamwise velocity at $z=2H$ was chosen as the reference velocity $U_\mathrm{ref} = \qty{1.5}{\meter\per\second}$ to normalize the velocity samples.
The Reynolds number based on the reference velocity and block height was approximately $Re_H=9600$.
The concentration samples were normalized by the reference concentration
\begin{equation}
    C_\mathrm{ref} = \frac{Q_C}{U_\mathrm{ref}HL},
    \label{eq:numsim:2}
\end{equation}
where $Q_C = \qty{100}{\ppm\,\meter^3\per\second}$ is the total pollutant emission rate, and $L$ is the length of the line source, which is equal to the span of the simulation domain (2.5$H$).
The snapshots were sampled after a  simulation time of $1300 H/U_\mathrm{ref}$, to ensure that the boundary layer has developed and the flow has reached a quasi-steady state.
The time signals for both the quantities (velocities and concentration) were recorded over a duration of \qty{80}{\second} ($\approx 1200 H/U_\mathrm{ref}$) at a sampling frequency of $f_s = \qty{1000}{\hertz}$ ($\Delta t = 0.015 H/U_\mathrm{ref}$), providing $N_t = 80000$ snapshots.

The flow in the canyon conforms to the skimming flow regime.
It is characterized by a principal recirculating cell with smaller counter-rotating vortices near the corners, as shown by the streamlines in \fref{fig:numsim:1b}.
The mean concentration is plotted in \fref{fig:numsim:1c}.
The emission follows the recirculation in the canyon and travels along the leeward wall before most of it re-enters the canyon upon reaching the roof level rather than being extracted and transported by the external flow. 
The simulation was also  validated in terms of turbulent quantities (mean velocities and turbulent kinetic energy) in \citet{combettePODGalerkinReducedorder2026} by comparing the solutions with reference results from the literature.

It has been shown that the pollutant removal from the canyon is mainly driven by turbulent fluctuations \citep{michiokaLargeEddySimulationMechanism2011}.
These fluctuations are characterized by complex multiscale and chaotic motions.
In order to perform meaningful dimensionality reduction, the flow must therefore be decomposed into fluctuation patterns belonging to specific scales, and subsequently limited to a \emph{reduced space}, \ie a finite set of most energetic coherent flow structures.
For the statistically stationary canyon flow, SPOD is considered in this work to extract the spatiotemporal structures and their expansion coefficients to describe the temporal dynamics.
In the next section, the reduced-order model approach based on the SPOD coefficients for predicting the canyon flow dynamics is presented.

%% file: section-method.tex
% ---------------------------------------------------------------------------
\section{Data-driven reduced-order modeling and prediction}
% ---------------------------------------------------------------------------
\label{sec:method}

This section details the mathematical formulation and implementation of the proposed data-driven framework.
The architecture is modular, designed to decouple the extraction of coherent spatial structures from the modeling of their temporal dynamics.
The methodology consists of four sequential stages:
(i) the decomposition of the high-fidelity velocity field into spectrally coherent modes using SPOD (\sref{sec:method:spod});
(ii) the nonlinear compression of the resulting modal coefficients into a compact latent space via an autoencoder (\sref{sec:method:ae});
(iii) the temporal forecasting of this latent state using a recurrent LSTM network (\sref{sec:method:lstm}); and
(iv) the reconstruction of the pollutant concentration field from the predicted flow field via a CNN (\sref{sec:method:cnn}).
The integration of these modules into a cohesive prediction pipeline is summarized in \sref{sec:method:rom}.

% ---------------------------------------------------------------------------
\subsection{SPOD modes and time-domain coefficients}
% ---------------------------------------------------------------------------
\label{sec:method:spod}

Let the vector $\mathbf{q}_k \inR{N_{xv}}$ represent the instantaneous state of a flow field variable $\mathbf{q}(\mathbf{x}, t)$ at time $t_k$ on a discrete set of points in the spatial domain $\Omega$. 
The length of the vector is given as $N_{xv} = N_x \times N_v$, where $N_x$ is the number of grid points and $N_v$ is the number of flow variables.
From the snapshot data available at $N_t$ equally spaced time instances $t_k = k\Delta t$, for $k=0,1,\ldots,N_t-1$ in the range $0\leq t \leq T$, the snapshot matrix $\mathbf{Q}$ can be compiled as
\begin{equation}
    \mathbf{Q} = \begin{bmatrix}
    \mid & \mid &  & \mid \\
    \mathbf{q}_1 & \mathbf{q}_2 & \dots & \mathbf{q}_{N_t} \\
    \mid & \mid &  & \mid
    \end{bmatrix} \inR{N_{xv} \times N_t}.
    \label{eq:method:spod:1}
\end{equation}
For the canyon flow described in \sref{sec:numsim}, the vector $\mathbf{q}_k$ represents the velocity fluctuations in the streamwise $u^\prime$ and vertical $w^\prime$ directions, implying $N_v=2$.
The velocity fluctuations are computed by decomposing the velocity components as 
\begin{subequations}
\label{eq:method:spod:2}
\begin{align}
    u^\prime(\mathbf{x},t) &= u(\mathbf{x},t) - \mathbb{E}(u(\mathbf{x},t)), 
    \label{eq:method:spod:2a} \\
    w^\prime(\mathbf{x},t) &= w(\mathbf{x},t) - \mathbb{E}(w(\mathbf{x},t)),
    \label{eq:method:spod:2b}
\end{align}
\end{subequations}
where $\mathbb{E}(\cdot)$ denotes expectation.
The snapshot vector is assembled by stacking the two velocity components as
\begin{equation}
    \mathbf{q}_k = \left[ u^\prime_k(\mathbf{x}_1), u^\prime_k(\mathbf{x}_2), \ldots, u^\prime_k(\mathbf{x}_{N_x}), w^\prime_k(\mathbf{x}_1), w^\prime_k(\mathbf{x}_2), \ldots, w^\prime_k(\mathbf{x}_{N_x}) \right]^\tr.
    \label{eq:method:spod:3}
\end{equation}

SPOD seeks an expansion of the flow variable in the frequency domain of the form given as
\begin{equation}
    \mathbf{q}(\mathbf{x},t) = \int_{-\infty}^{+\infty}\sum_{n=1}^\infty a_n(f) \mathbf{\Phi}_n(\mathbf{x},f) \mathrm{e}^{\mathrm{i}2\pi ft} \diff f,
    \label{eq:method:spod:4}
\end{equation}
where $f$ is the frequency, $\mathbf{\Phi}_n(\mathbf{x},f)$ are the SPOD modes, and $a_n(f)$ are the corresponding mode coefficient.
Applying Fourier transform on both sides of \eref{eq:method:spod:4} gives
\begin{equation}
    \widehat{\mathbf{q}}(\mathbf{x},f) = \sum_{n=1}^\infty a_n(f) \mathbf{\Phi}_n(\mathbf{x},f),
    \label{eq:method:spod:5}
\end{equation}
where $\widehat{(\cdot)}$ denotes the Fourier transform in time.
It has been shown that the modes can be obtained as a solution to the eigenvalue problem \citep{towneSpectralProperOrthogonal2018} given as
\begin{equation}
    \int_\Omega \mathbf{S}(\mathbf{x}, \mathbf{x}^\prime, f) \mathbf{\mathcal{W}}(\mathbf{x}^\prime) \mathbf{\Phi}(\mathbf{x}^\prime,f) \diff\mathbf{x}^\prime = \lambda(f) \mathbf{\Phi}(\mathbf{x},f),
    \label{eq:method:spod:6}
\end{equation}
where $\mathbf{\mathcal{W}}(\mathbf{x})$ accounts for the spatial weighting introduced in the definition of the inner product and $\mathbf{S}(\mathbf{x}, \mathbf{x}^\prime, f)$ is the Fourier transformed two-point space–time correlation tensor, also known as the cross-spectral density (CSD) tensor.
To obtain converged estimates of the spectral densities, it is necessary to appropriately average the spectra over multiple realizations of the flow. 
This can be accomplished using Welch’s method as outlined in \cite{towneSpectralProperOrthogonal2018}.
The data matrix $\mathbf{Q}$ is first partitioned into $N_\mathrm{blk}$ overlapping blocks of length $N_\mathrm{fft}$
\begin{equation}
    \mathbf{Q}^{(i)} = \begin{bmatrix}
    \mid & \mid &  & \mid \\
    \mathbf{q}^{(i)}_1 & \mathbf{q}^{(i)}_2 & \dots & \mathbf{q}^{(i)}_{N_\mathrm{fft}} \\
    \mid & \mid &  & \mid
    \end{bmatrix} \inR{N_{xv} \times N_\mathrm{fft}} 
    \quad \forall\, i = 1,\ldots,N_\mathrm{blk}.
    \label{eq:method:spod:7}
\end{equation}
The $k$-th column in the $i$-th block corresponds to a column in the original data matrix \eref{eq:method:spod:1}
\begin{equation}
    \mathbf{q}_{k}^{(i)} = \mathbf{q}_{k+(i-1)(N_\mathrm{fft} - N_\mathrm{ovlp})} \inR{N_{xv}} 
    \quad \forall\, i = 1,\ldots,N_\mathrm{blk} \, \text{ and } \, k = 1,\ldots,N_\mathrm{fft},
    \label{eq:method:spod:8}
\end{equation}
where $N_\mathrm{ovlp}$ is the number of snapshots by which the blocks overlap.
Assuming ergodicity, each block is regarded as equally representative of the whole data.
In practice, temporal discrete Fourier transform (DFT) is applied to \eref{eq:method:spod:7} to obtain the Fourier-transformed data matrix
\begin{equation}
    \widehat{\mathbf{Q}}^{(i)} = \begin{bmatrix}
    \mid & \mid &  & \mid \\
    \widehat{\mathbf{q}}^{(i)}_1 & \widehat{\mathbf{q}}^{(i)}_2 & \dots & \widehat{\mathbf{q}}^{(i)}_{N_\mathrm{fft}} \\
    \mid & \mid &  & \mid
    \end{bmatrix} \inC{N_{xv} \times N_\mathrm{fft}} 
    \quad \forall\, i = 1,\ldots,N_\mathrm{blk},
    \label{eq:method:spod:9}
\end{equation}
where $\widehat{\mathbf{q}}_{k}^{(i)} \inC{N_{xv}}$ is the Fourier realization of the data at the discrete frequency $f_k$ in the $i$-th block.
The resolved frequencies are
\begin{equation}
    f_k = \begin{cases}
      \dfrac{k-1}{N_\mathrm{fft} \Delta t} & \quad \text{for}\, k \leq N_\mathrm{fft}/2 \\
      \dfrac{k-1-N_\mathrm{fft}}{N_\mathrm{fft} \Delta t} &  \quad \text{for}\, k > N_\mathrm{fft}/2
    \end{cases}, \quad\, \forall\, k=1,\ldots,N_\mathrm{fft}.
    \label{eq:method:spod:9-nfft}
\end{equation}
When the original data \eref{eq:method:spod:1} is real, the transformed data in \eref{eq:method:spod:9} at negative frequencies $f_k$ corresponds to the conjugates of the corresponding positive frequencies.
The redundant negative frequencies are therefore omitted from the SPOD feature space, and a reduced number of frequencies $N_\mathrm{fc} = \ceil{N_\mathrm{fft}/2} + 1$ is considered, where $\ceil{\cdot}$ is the ceil operator.

The Fourier realizations are reorganized with respect to frequencies.
The matrix containing all the realizations at the frequency $f_k$ from each block is given as
\begin{equation}
    \widehat{\mathbf{Q}}_{f_k} = 
    \begin{bmatrix}
    \mid & \mid &  & \mid \\
    \widehat{\mathbf{q}}_{k}^{(1)} & \widehat{\mathbf{q}}_{k}^{(2)} & \dots & \widehat{\mathbf{q}}_{k}^{(N_\mathrm{blk})} \\
    \mid & \mid &  & \mid
    \end{bmatrix} \inC{N_{xv} \times N_\mathrm{blk}} 
    \quad \forall\, k = 1,2,\ldots,N_\mathrm{fc}.
    \label{eq:method:spod:10}
\end{equation}
The rearranged Fourier-transformed data matrix is used to calculate the CSD tensor corresponding to the frequency $f_k$
\begin{equation}
    \mathbf{S}_{f_k} = \frac{1}{N_\mathrm{blk}-1} \widehat{\mathbf{Q}}_{f_k} \widehat{\mathbf{Q}}_{f_k} ^\ast \inC{N_{xv} \times N_{xv}},
    \label{eq:method:spod:11}
\end{equation}
where $N_\mathrm{blk}-1$ is a normalization factor applied for data centered about the sample mean.
The eigenvectors and eigenvalues of the tensor $\mathbf{S}_{f_k}$ are the SPOD modes and the associated energies, respectively.

In general, the number of spatial grid points is much larger than the number of realizations, \ie $N_{xv} \gg N_\mathrm{blk}$. It is then computationally more tractable to solve the analogous eigenvalue problem, given as
\begin{equation}
    \widehat{\mathbf{Q}}_{f_k}^\ast \mathbf{W} \widehat{\mathbf{Q}}_{f_k} \mathbf{\mathbf{\Psi}}_{f_k} = \mathbf{\mathbf{\Psi}}_{f_k} \mathbf{\Lambda}_{f_k},
    \label{eq:method:spod:12}
\end{equation}
where $\mathbf{W}$ accounts for both the weight $\mathbf{\mathcal{W}}(\mathbf{x})$ and the numerical quadrature of the integral in \eqref{eq:method:spod:6}.
The SPOD modes at frequency $f_k$ are recovered as
\begin{equation}
    \mathbf{\Phi}_{f_k} = \widehat{\mathbf{Q}}_{f_k} \mathbf{\Psi}_{f_k} \mathbf{\Lambda}_{f_k}^{-1/2} \inC{N_{xv} \times N_\mathrm{blk}}.
    \label{eq:method:spod:13}
\end{equation}
The matrices 
$\mathbf{\Lambda}_{f_k} = \diag(\lambda_{f_k}^{(1)}, \lambda_{f_k}^{(2)}, \ldots, \lambda_{f_k}^{(N_\mathrm{blk})})$ 
and 
$\mathbf{\Phi}_{f_k} = [\boldsymbol{\phi}_{f_k}^{(1)}, \boldsymbol{\phi}_{f_k}^{(2)}, \ldots, \boldsymbol{\phi}_{f_k}^{(N_\mathrm{blk})}]$ 
contain the SPOD energies and modes respectively. 
By construction, at any given frequency $f_k$, the modes are orthonormal in the spatial inner product, \ie $\mathbf{\Phi}_{f_k}^\ast \mathbf{W} \mathbf{\Phi}_{f_k} = \mathbf{I}_{N_\mathrm{blk}}$.
The matrix of SPOD modes $\mathbf{\Phi} \inC{N_{xv} \times (N_\mathrm{blk} \times N_\mathrm{fc})}$ is assembled by grouping the modes with respect to frequency
\begin{equation}
\begin{split}
    \mathbf{\Phi} & = [ \mathbf{\Phi}_{f_1}, \mathbf{\Phi}_{f_2}, \ldots, \mathbf{\Phi}_{f_{N_\mathrm{fc}}} ] \\ 
    & = [ \underbrace{\boldsymbol{\phi}_{f_1}^{(1)}, \boldsymbol{\phi}_{f_1}^{(2)}, \ldots, \boldsymbol{\phi}_{f_1}^{(N_\mathrm{blk})}}_{\mathbf{\Phi}_{f_1}}, \underbrace{\boldsymbol{\phi}_{f_2}^{(1)}, \boldsymbol{\phi}_{f_2}^{(2)}, \ldots, \boldsymbol{\phi}_{f_2}^{(N_\mathrm{blk})}}_{\mathbf{\Phi}_{f_2}}, \ldots, \underbrace{\boldsymbol{\phi}_{f_{N_\mathrm{fc}}}^{(1)}, \boldsymbol{\phi}_{f_{N_\mathrm{fc}}}^{(2)}, \ldots, \boldsymbol{\phi}_{f_{N_\mathrm{fc}}}^{(N_\mathrm{blk})}}_{\mathbf{\Phi}_{f_{N_\mathrm{fc}}}} ].
    \label{eq:method:spod:14}
    \end{split}
\end{equation}
Each SPOD mode is associated with an expansion coefficient, which is discussed in \sref{sec:method:spod:coeff}.
In order to ensure that the SPOD modes are well-resolved, $N_\mathrm{blk}$ and $N_\mathrm{fc}$ must be sufficiently large, which leads to a high-dimensional SPOD space.
However, for reduced-order modeling application, this can be computationally intractable. 
This therefore necessitates the implementation of dimensionality reduction techniques, as detailed in \sref{sec:method:spod:dim}.

% ---------------------------------------------------------------------------
\subsubsection{Time-domain coefficients}
% ---------------------------------------------------------------------------
\label{sec:method:spod:coeff}

In this work, a reduced-order model (ROM) is formulated in terms of expansion coefficients obtained from SPOD modes.
A projection-based approach outlined in \cite{nekkantiFrequencyTimeAnalysis2021} is used to obtain the coefficients in the time domain.
The matrix of expansion coefficients is obtained by weighted oblique projection of the original data onto the SPOD modes
\begin{equation}
    \mathbf{A} = \left( \mathbf{\Phi}^\ast\mathbf{W}\mathbf{\Phi} \right)^{-1} \mathbf{\Phi}^\ast \mathbf{W} \mathbf{Q} \inC{(N_\mathrm{blk} \times N_\mathrm{fc}) \times N_t}.
    \label{eq:method:spod:coeff:1}
\end{equation}
An oblique projection is needed as the inter-frequency SPOD modes are not orthogonal with respect to the spatial inner product, \ie 
$\mathbf{\Phi}_{f_{k^\prime}}^\ast \mathbf{W} \mathbf{\Phi}_{f_k} \neq \mathbf{I}_{N_\mathrm{blk}}$ 
if $k^\prime \neq k$.
The flow field variable can be calculated from the SPOD mode and expansion coefficient as
\begin{equation}
    \widetilde{\mathbf{Q}} = \mathbf{\Phi} \mathbf{A},
    \label{eq:method:spod:coeff:2}
\end{equation}
where $\widetilde{\mathbf{Q}}$ is an approximation of the original data matrix $\mathbf{Q}$.
The individual field variable vector $\widetilde{\mathbf{q}}_k$ is represented as
\begin{equation}
    \widetilde{\mathbf{q}}_k = \mathbf{\Phi} \left( \mathbf{\Phi}^\ast\mathbf{W}\mathbf{\Phi} \right)^{-1} \mathbf{\Phi}^\ast \mathbf{W} \mathbf{q}_k.
    \label{eq:method:spod:coeff:3}
\end{equation}
From the perspective of recovering the overall energy of the field, summing the eigenvalues over all the modes and integrating over all the frequencies gives 
\begin{equation}
    2\overline{k} = \int_{-\infty}^{+\infty}\sum_n \lambda^{(n)}(f) \diff f,
    \label{eq:method:spod:coeff:4}
\end{equation}
where $\overline{k}$ is the spatial average TKE.
It can be inferred that if only a subset of the full space of modes and frequencies is used, the dimension of the feature space $\mathbf{\Phi}$ would be reduced at the cost of the overall TKE represented by the preserved modal energies.
This consequently reduces the dimension of $\mathbf{A}$.
The criteria used for selecting the relevant features for dimensionality reduction are detailed next.

% ---------------------------------------------------------------------------
\subsubsection{Dimensionality reduction of SPOD space}
% ---------------------------------------------------------------------------
\label{sec:method:spod:dim}

In fluid flow problems, the number of features in the SPOD space ($N_\mathrm{fc} \times N_\mathrm{blk}$) can be sufficiently large to render it unsuitable for application in model order reduction.
To alleviate this bottleneck, two strategies are suggested to reduce the dimension of mode space:
\begin{enumerate}
    \item The SPOD space is truncated by discarding the features which do not contribute significantly to the total energy. In \citet{schmidtGuideSpectralProper2020}, the dominant mechanisms are characterized by a large separation between the eigenvalues associated with the first and second SPOD modes $\Delta\lambda_{f_k} = \abs{\lambda_{f_k}^{(1)} - \lambda_{f_k}^{(2)}}$.
    When this criterion is verified, it indicates that the mechanism associated with the leading mode is physically dominant. 
    To determine the $N_f$ physically dominant frequencies to be retained in the SPOD modes, we first sort the values $\Delta\lambda_{f_k}$ in descending order. We thus define
    $$
    \mathcal{S}_N = \{\nu_1, \nu_2,\cdots,\nu_N \mid 
    \Delta\lambda_{\nu_1} \geq \Delta\lambda_{\nu_2} \geq \ldots \geq \Delta\lambda_{\nu_{N}}\},
    $$
    where the values of $\nu_i$ ($i=1,\cdots,N$) correspond to frequencies $f_k$.
    
    The relative information content (RIC) is typically used as a metric of contribution to the total energy. In the context of SPOD, this metric is defined as
    \begin{equation}
        \text{RIC}(N) =
        \frac{\displaystyle\sum_{\nu_i\in\mathcal{S}_N}\sum_{n=1}^{N_\mathrm{blk}} \lambda_{\nu_i}^{(n)} }{\displaystyle\sum_{\nu_i\in\mathcal{S}_{N_\mathrm{fc}}}\sum_{n=1}^{N_\mathrm{blk}} \lambda_{\nu_i}^{(n)}}.
        \label{eq:method:spod:dim:1}
    \end{equation}    
    
    In practice, a threshold value $\varepsilon_\text{RIC} \in (0,1]$ can be assigned in order to obtain the number of leading frequencies $N_f$ such that $\text{RIC}(N_f) \geq \varepsilon_\text{RIC}$.
    The value of the threshold $N_f$  therefore dictates the size of the feature space. We have $N_f \to N_\mathrm{fc}$ as $\varepsilon_\text{RIC} \to 1$.

    \item The SPOD modes $\mathbf{\Phi}_{f_k}$ corresponding to different values of frequency $f_k$ exhibit spatial similarity which can be leveraged to further reduce the dimension of the feature space.
    In the current context, \emph{similarity} refers to the observation that the most energetic modes within a small frequency range depict similar large-scale flow structures, albeit with some differences in small-scale features.
    Moreover, some modes, especially the ones at high frequencies, change gradually with the frequency, and therefore exhibit similar shapes.
    For a quantitative measure of the similarity, \cite{zhangIdentificationThreedimensionalFlow2021} introduced the coefficient $\gamma$, defined as
    \begin{equation}
        \gamma_{(k,n), (k^\prime,n^\prime)} = \sqrt{\frac{\dotp{\boldsymbol{\phi}_{f_k}^{(n)}}{\boldsymbol{\phi}_{f_{k^\prime}}^{(n^\prime)}}{\Omega} \dotp{\boldsymbol{\phi}_{f_k}^{(n)}}{\boldsymbol{\phi}_{f_{k^\prime}}^{(n^\prime)}}{\Omega}^\ast}{\dotp{\boldsymbol{\phi}_{f_k}^{(n)}}{\boldsymbol{\phi}_{f_k}^{(n)}}{\Omega} \dotp{\boldsymbol{\phi}_{f_{k^\prime}}^{(n^\prime)}}{\boldsymbol{\phi}_{f_{k^\prime}}^{(n^\prime)}}{\Omega}^\ast}}, \quad \text{with} \quad 0 \leq \gamma_{(k,n), (k^\prime,n^\prime)} \leq 1,
        \label{eq:method:spod:dim:2}
    \end{equation}
    where $\dotp{\cdot}{\cdot}{\Omega}$ denotes the inner product over the domain $\Omega$. 
    Owing to the orthogonality property of SPOD modes, the denominator is equal to 1, and \eref{eq:method:spod:dim:2} can be simplified as 
    \begin{equation}
        \gamma_{(k,n), (k^\prime,n^\prime)} = \abs{\dotp{\boldsymbol{\phi}_{f_k}^{(n)}}{\boldsymbol{\phi}_{f_{k^\prime}}^{(n^\prime)}}{\Omega}},
        \label{eq:method:spod:dim:3}
    \end{equation}
    where $\abs{\cdot}$ denotes the modulus of a complex number.
    A threshold value $\varepsilon_\gamma \in [0,1]$ can be assigned to identify the modes which exhibit similar features.
    The reduction of the dimension of feature space is achieved by only preserving the modes with $\gamma \leq \varepsilon_\gamma$ and discarding the rest, ensuring that the preserved modes represent unique structures.
\end{enumerate}

In the following steps, the two criteria described above are applied sequentially to reduce the dimension of the SPOD space.
The number of modes preserved after the dimensionality reduction is represented as $N_m$.
In practice, when the criteria are applied to turbulent canyon flow, $N_m$ can be large which can render the modal space unsuitable for model order reduction.
In order to obtain a more compact feature space, an additional data compression step needs to be performed.
Unlike the RIC- and similarity-based criteria, this dimensionality reduction step is not interpretable but instead ensures that the framework scales over a large range of feature space dimension.
In this work, a neural network based approach is proposed to generate a low-dimensional space, as discussed next.

% ---------------------------------------------------------------------------
\subsection{Autoencoder (AE) latent space}
% ---------------------------------------------------------------------------
\label{sec:method:ae}

\begin{figure}
    \centering
    \begin{subfigure}{0.45\textwidth}
        \centering
        \includegraphics[width=0.82\textwidth]{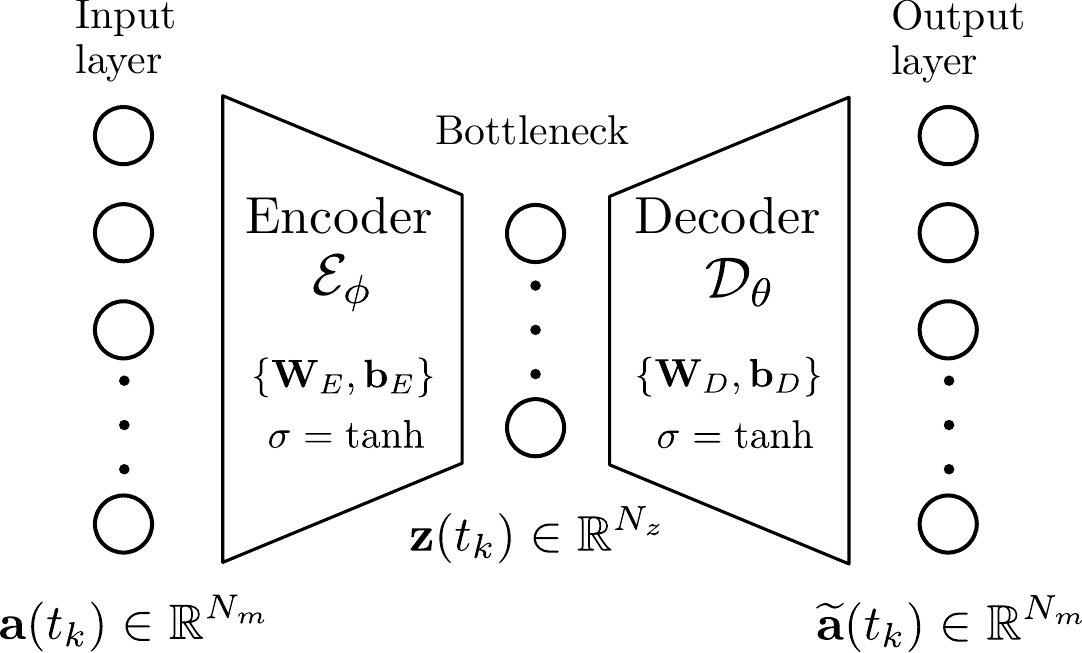}
        \caption{}
        \label{fig:method:arch:1a}
    \end{subfigure}
    \hspace{1em}
    \begin{subfigure}{0.45\textwidth}
        \centering
        \includegraphics[width=0.7\textwidth]{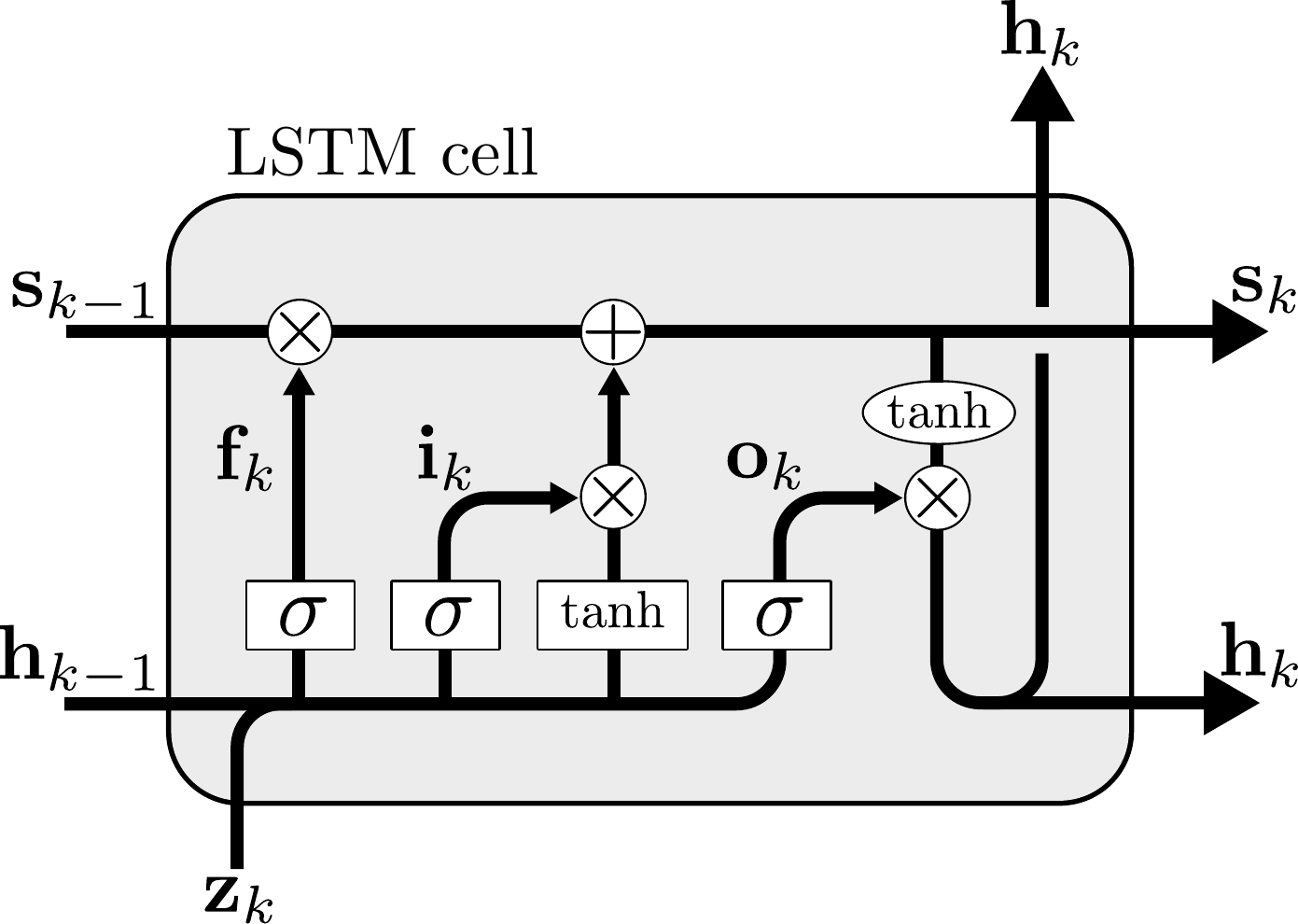}
        \caption{}
        \label{fig:method:arch:1b}
    \end{subfigure}
    \\
    \begin{subfigure}{0.6\textwidth}
        \centering
        \includegraphics[width=0.7\textwidth]{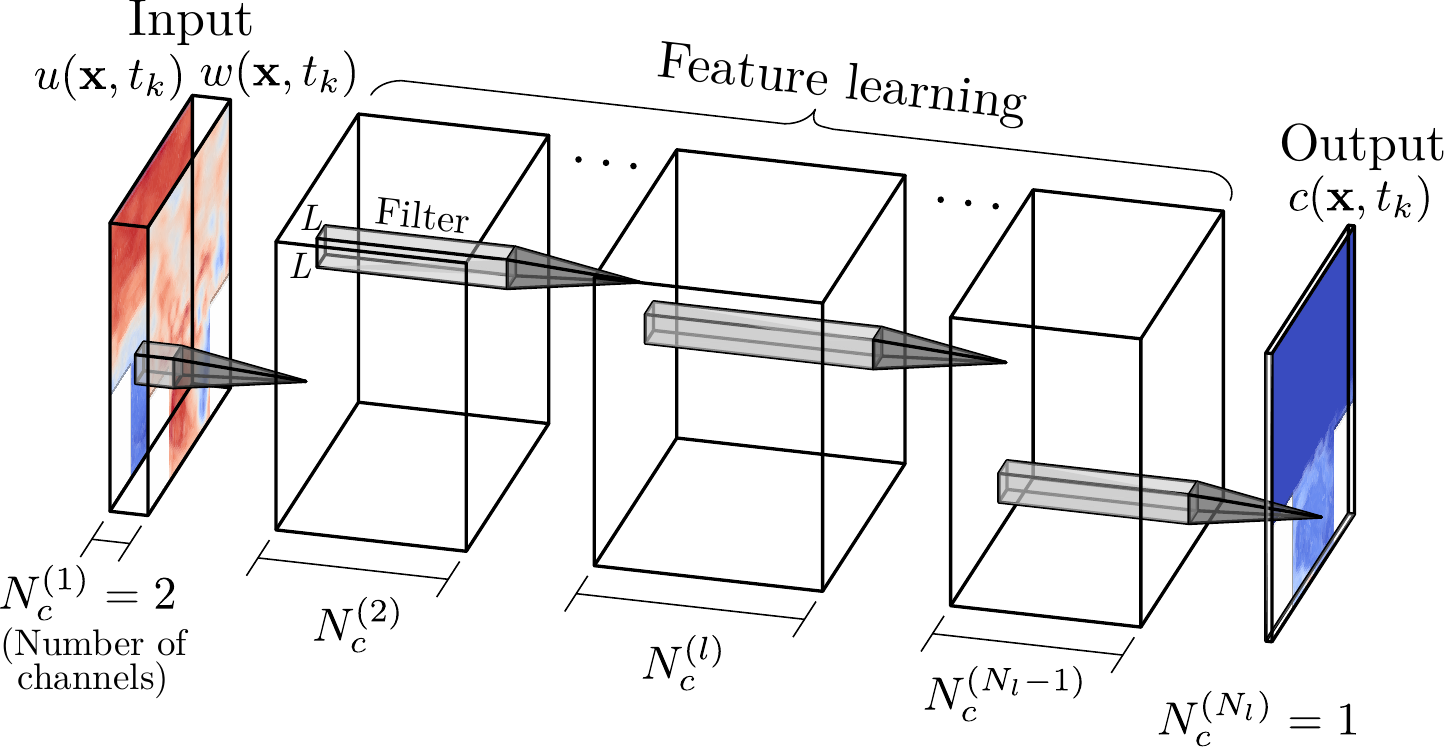}
        \caption{}
        \label{fig:method:arch:1c}
    \end{subfigure}
    \caption{
        Schematic of the neural network architectures used in the ROM framework: (a) Autoencoder, (b) Long Short-Term Memory (LSTM) cell, and (c) Convolutional Neural Network (CNN).
    }
    \label{fig:method:arch:1}
\end{figure}

The autoencoder (AE) is a class of neural network architecture designed to achieve dimensionality reduction or feature learning for high-dimensional input data.
An efficient representation of the input data, referred to as the \emph{latent space} or \emph{embedding}, is learned through an unsupervised training process.
Structurally, an AE consists of three principal components: an encoder, a bottleneck (representing the latent space), and a decoder.
Nonlinear mappings are applied by the encoder to compress the input vector into the lower-dimensional bottleneck layer.
Subsequently, nonlinear mappings are applied by the decoder to reconstruct the original input vector from this latent embedding.
During the training stage, the latent space is forced to capture the most salient features of the input while discarding the less significant information, effectively acting as a nonlinear filter.

In this work, the AE is trained on the time-domain coefficients $\mathbf{A} \inC{N_m \times N_t}$ corresponding to the preserved SPOD modes (see \sref{sec:method:spod:coeff}).
The primary objective is to derive a low-dimensional representation of these coefficients.
Since standard neural networks operate on real-valued data, the complex-valued coefficients are treated by separating their real and imaginary components.
Two separate AEs are trained, \ie one for the real component $\Re(\mathbf{A})$ and one for the imaginary component $\Im(\mathbf{A})$.
The resulting latent representations are then combined to form a complex latent vector $\mathbf{z}(t) \inC{N_z}$, or $\mathbf{Z} \inC{N_z \times N_t}$ in matrix form, where typically $N_z \ll N_m$.

The encoder function, denoted as $\mathcal{E}_\phi: \mathbb{R}^{N_m} \rightarrow \mathbb{R}^{N_z}$ and parameterized by $\phi$, maps the input component to the latent space.
The decoder function, denoted as $\mathcal{D}_\theta: \mathbb{R}^{N_z} \rightarrow \mathbb{R}^{N_m}$ and parameterized by $\theta$, generates the reconstructed output $\widetilde{\mathbf{A}}$ from the compressed latent space.
The complete autoencoder operation is defined as the composition of the encoder and decoder functions,
\begin{equation}
    \widetilde{\mathbf{A}} = (\mathcal{D}_\theta \circ \mathcal{E}_\phi)\mathbf{A},
    \label{eq:method:ae:1}
\end{equation}
where $(\mathcal{D}_\theta \circ \mathcal{E}_\phi)\mathbf{A} = \mathcal{D}_\theta(\mathcal{E}_\phi(\mathbf{A}))$.
A schematic of the AE network is presented in \fref{fig:method:arch:1a}.
The learning objective is to minimize the reconstruction error between the input $\mathbf{A}$ and the output $\widetilde{\mathbf{A}}$.
This error is quantified by a loss function defined as the mean squared error (MSE),
\begin{equation}
    \phi^\ast, \theta^\ast = \argmin_{\phi, \theta} \mathcal{L}(\phi, \theta), \quad \text{where} \quad
    \mathcal{L}(\phi, \theta) = \frac{1}{N_t} \sum_{k=1}^{N_t} \norm{\mathbf{a}(t_k) - (\mathcal{D}_\theta \circ \mathcal{E}_\phi)\mathbf{a}(t_k) }_2^2.
    \label{eq:method:ae:2}
\end{equation}

The network consists of dense (fully connected) layers, where every neuron in layer $l$ is connected to every neuron in layer $l+1$.
The output of the $l$-th layer for the encoder and decoder is expressed as
\begin{subequations}
\label{eq:method:ae:3}
\begin{align}
    \mathbf{x}_E^{(l)} &= \sigma_E \left( \mathbf{W}_E^{(l)} \mathbf{x}_E^{(l-1)} + \mathbf{b}_E^{(l)} \right) \inR{n_E^{(l)}}, 
    \quad \forall\, l = 1,2,\ldots,N_{l,E},
    \label{eq:method:ae:3a} \\
    \mathbf{x}_D^{(l)} &= \sigma_D \left( \mathbf{W}_D^{(l)} \mathbf{x}_D^{(l-1)} + \mathbf{b}_D^{(l)} \right) \inR{n_D^{(l)}}, 
    \quad \forall\, l = 1,2,\ldots,N_{l,D},
    \label{eq:method:ae:3b}
\end{align}
\end{subequations}
where the subscripts $E$ and $D$ denote the encoder and decoder, respectively.
Here, $\sigma$ represents element-wise activation functions, $\mathbf{W}^{(l)}$ are the weight matrices, $\mathbf{b}^{(l)}$ are the bias vectors, $n^{(l)}$ is the number of neurons in layer $l$, and $N_{l}$ denotes the number of layers.
Since the SPOD coefficients are zero-centered and fluctuate between positive and negative values, the hyperbolic tangent function, $\sigma(x) = \tanh(x)$, is selected as the activation function for the hidden layers.
The input to the encoder is the vector of SPOD coefficients, $\mathbf{x}_E^{(0)} = \mathbf{a}(t_k) \inR{N_m}$, while the input to the decoder is the latent vector, $\mathbf{x}_D^{(0)} = \mathbf{z}(t_k) \inR{N_z}$.

To ensure the output can span the full dynamic range of the coefficients, linear transformations are employed for the final output layers of both the encoder (the bottleneck) and the decoder,
\begin{subequations}
\label{eq:method:ae:4}
\begin{align}
\mathbf{z}(t_k) &= \mathbf{W}_E^{(N_{l,E})} \mathbf{x}_E^{(N_{l,E}-1)} + \mathbf{b}_E^{(N_{l,E})},
    \label{eq:method:ae:4a} \\
    \widetilde{\mathbf{a}}(t_k) &= \mathbf{W}_D^{(N_{l,D})} \mathbf{x}_D^{(N_{l,D}-1)} + \mathbf{b}_D^{(N_{l,D})},
    \label{eq:method:ae:4b}
\end{align}
\end{subequations}
where $k=1,2,\ldots,N_t$.

The trainable parameters of the encoder, $\phi = \{\mathbf{W}_E, \mathbf{b}_E \}$, and the decoder, $\theta = \{\mathbf{W}_D, \mathbf{b}_D \}$, are initialized at the start of training.
These parameters are updated iteratively using stochastic gradient descent (SGD) based on random subsets of the data, known as minibatches.
The parameter update rule between iterations $k$ and $k+1$ is given as 
\begin{equation}
    \phi_{k+1} = \phi_{k} - \eta \nabla_\phi \mathcal{L}, \quad \text{and} \quad
    \theta_{k+1} = \theta_{k} - \eta \nabla_\theta \mathcal{L},
    \label{eq:method:ae:5}
\end{equation}
where $\eta$ is the learning rate and $\nabla$ is the gradient operator computed using backpropagation.
The Adam optimization algorithm is employed to adaptively adjust the learning rates based on the first and second moments of the gradients.
To monitor convergence and prevent overfitting, the loss is evaluated on a validation dataset that is excluded from the training process.
Training is halted when the validation error stabilizes, ensuring the model generalizes well to unseen data.

Once the compact latent space $\mathbf{z}(t)$ is obtained, its temporal evolution is modeled using a recurrent neural network, as detailed in the following section.

% ---------------------------------------------------------------------------
\subsection{Long short-term memory (LSTM) time forecasting}
% ---------------------------------------------------------------------------
\label{sec:method:lstm}

LSTM network is a type of recurrent neural network capable of learning long-term dependencies.
The \emph{recurrent} nature of the network is derived from the fact that neuron outputs from each time step in the prediction are utilized to influence the predictions in the future time steps. 
Such an embedding allows the information to persist and learn the long and short-term correlations in time which is critical for forecasting application.
Given a driving sequence of latent space variables $\mathbf{z}(t)$, the objective of LSTM is to learn the mapping
\begin{equation}
    \mathbf{h}(t_k) = f(\mathbf{h}(t_{k-1}), \mathbf{z}(t_k)),
    \label{eq:method:lstm:1}
\end{equation}
where $\mathbf{h}(t_k) \inR{N_h}$ is the hidden state of size $N_h$ of the cell at the current time step.
The schematic of LSTM cell is shown in \fref{fig:method:arch:1b}.
Each cell also has an associated cell state $\mathbf{s}_k$ at time step $t_k$ which is updated through transformations known as \emph{gates}.
Three gates control the access to the memory cell: the \emph{input gate} which adds useful information to the cell, the \emph{forget gate} which discards the information that is no longer useful in the cell state, and the \emph{output gate} which decides the part of the cell state that can be forwarded as hidden state for the time step.
These are formulated as follows
\begin{subequations}
    \label{eq:method:lstm:2}
    \begin{align}
        \text{Input gate} : \mathbf{i}_k &= \sigma(\mathbf{W}_i[\mathbf{h}_{k-1};\mathbf{z}_k] + \mathbf{b}_i), 
        \label{eq:method:lstm:2a} \\
        \text{Forget gate} : \mathbf{f}_k &= \sigma(\mathbf{W}_f[\mathbf{h}_{k-1};\mathbf{z}_k] + \mathbf{b}_f), 
        \label{eq:method:lstm:2b} \\
        \text{Output gate} : \mathbf{o}_k &= \sigma(\mathbf{W}_o[\mathbf{h}_{k-1};\mathbf{z}_k] + \mathbf{b}_o),
        \label{eq:method:lstm:2c}
    \end{align}
\end{subequations}
and the cell state and the hidden states are updated as
\begin{subequations}
    \label{eq:method:lstm:3}
    \begin{gather}
        \mathbf{s}_k = \mathbf{f}_k \odot \mathbf{s}_{k-1} + \mathbf{i}_k \odot \tanh(\mathbf{W}_s[\mathbf{h}_{k-1};\mathbf{z}_k] + \mathbf{b}_s), 
        \label{eq:method:lstm:3a} \\
        \mathbf{h}_k = \mathbf{o}_k \odot \tanh(\mathbf{s}_k).
        \label{eq:method:lstm:3b}
    \end{gather}
\end{subequations}
Here $[\mathbf{h}_{k-1};\mathbf{z}_k] \inR{N_h+N_z}$ is the concatenation of the previous hidden state $\mathbf{h}_{k-1} = \mathbf{h}(t_{k-1})$ and the current input $\mathbf{z}_k = \mathbf{z}(t_k)$, $\mathbf{W}_i, \mathbf{W}_f, \mathbf{W}_o, \mathbf{W}_s \inR{N_h \times (N_h+N_z)}$ and $\mathbf{b}_i, \mathbf{b}_f, \mathbf{b}_o, \mathbf{b}_s \inR{N_h}$ are the weights and biases for the corresponding gates, $\sigma(x) = 1/(1 + e^{-x})$ is the logistic sigmoid function, and $\odot$ represents element-wise multiplication.
The updated hidden state $\mathbf{h}_{k}$ can be passed to the next time step and can also be used to generate the output of the network.
Similarly to the previous section, the Adam optimizer is used to train the model.
In order to learn long-term time dependencies, the input and output batches can be formed to contain sequences of lengths $N_{t,\text{in}}$ and $N_{t,\text{out}}$ respectively.
The LSTM model \eref{eq:method:lstm:1} is modified to learn the mapping between the input and output sequences as
\begin{equation}
    \mathbf{h}(t_k, t_{k+1}, \ldots, t_{k+N_{t,\text{out}}-1}) = f(\mathbf{h}(t_{k-1}, t_{k-2}, \ldots, t_{k-N_{t,\text{in}}}), \mathbf{z}(t_{k-1}, t_{k-2}, \ldots, t_{k-N_{t,\text{in}}})).
    \label{eq:method:lstm:4}
\end{equation}
For autoregressive prediction of latent vector, where the model output at a previous time step $t_{k-1}$ is used as an input at the current time step $t_k$, it is common to have output for a single time step, \ie $N_{t,\text{out}} = 1$.

The predicted latent space variables are passed through the decoder $\mathcal{D}_\theta$ to obtain the predicted time-domain coefficients.
Using the SPOD modes that were retained in the procedure described in \sref{sec:method:spod:dim}, the predicted velocity field is obtained by the matrix multiplication \eref{eq:method:spod:coeff:2}.
In order to obtain the predicted concentration field from the velocity field, a neural network based mapping is performed, as discussed next.

% ---------------------------------------------------------------------------
\subsection{Mapping pollutant field via Convolutional Neural Network}
% ---------------------------------------------------------------------------
\label{sec:method:cnn}

In this numerical study, the pollutant concentration is treated as a passive scalar quantity, primarily driven by the advective transport of the velocity field.
In the skimming flow regime, pollutant dispersion is dominated by large-scale coherent structures.
Consequently, for a continuous emission source in a statistically stationary state, a strong spatial correlation is observed between the instantaneous velocity patterns (particularly with the canyon recirculation) and the scalar plume topology.
This correlation is leveraged to establish a direct, nonlinear mapping from the flow field to the scalar field, effectively bypassing the computational cost of solving the transport equation for every time step.

A convolutional neural network (CNN) is employed to approximate this mapping.
Unlike fully connected networks, the spatial correlation of the flow is preserved by CNNs, which are efficient at extracting local features, such as shear layers and stagnation points, that are important for dispersion.
Let $\widetilde{\mathbf{q}}_k \inR{N_x \times N_v}$ represent the reconstructed velocity field (comprising streamwise and vertical components; $N_v=2$) at time $t_k$, obtained from the SPOD-AE-LSTM pipeline.
The CNN, denoted as $\mathcal{F}_{CNN}$ and parameterized by weights $\theta_C$, is defined to map this input tensor to the estimated concentration field $\widetilde{\mathbf{c}}_k \inR{N_x \times 1}$:
\begin{equation}
    \widetilde{\mathbf{c}}_k = \mathcal{F}_\mathrm{CNN}( \widetilde{\mathbf{q}}_k; \theta_C ).
    \label{eq:method:cnn:1-1}
\end{equation}

A fully convolutional design is utilized for the architecture (illustrated in \fref{fig:method:arch:1c}).
Since the spatial resolution $N_x$ is preserved between the input and the output, the dimensionality reduction (pooling) or expansion (upsampling) operations typically found in classification networks are excluded from the current model.
Instead, the model relies on a sequence of convolutional layers to extract coherent spatial features through learnable filter operations.
The fundamental operation involves the summation of an element-wise product between a local patch of the input data and a filter kernel $\mathbf{w}$.

The output of the $n$-th channel of the $l$-th convolutional layer, denoted as $x^{(l)}_{ij,n}$ is mathematically expressed as
\begin{equation}
\begin{split}
    x^{(l)}_{ij,n} = \sigma \left( \sum_{m=1}^{N_{c,\text{in}}^{(l)}}\sum_{p=0}^{L-1}\sum_{q=0}^{L-1} w^{(l)}_{pq,mn} x^{(l-1)}_{(i+p)(j+q),m} + b^{(l)}_{n}\right),
    \quad \forall\, l = 1,2,\ldots,N_l \,\text{ and }\, n = 1,2,\ldots,N_{c,\text{out}}^{(l)},
    \label{eq:method:cnn:1}
    \end{split}
\end{equation}
where the indices $i,j$ denote the grid points along the streamwise and vertical dimensions, and $L$ represents the width and height of the filter kernel.
The weights $\mathbf{w}_{mn}\inR{L\times L}$ are shared between the $m$-th input channel and the $n$-th output channel. 
Here, $N_l$ is the total number of convolutional layers, $N_{c,\text{in}}^{(l)}$ and $N_{c,\text{out}}^{(l)}$ denote the number of input and output channels for layer $l$, $\mathbf{b}^{(l)}$ is the bias term, and $\sigma$ represents the rectified linear unit (ReLU) activation function.

As the input to the CNN consists of the streamwise and vertical velocity components, the first layer is initialized with $N_{c,\text{in}}^{(1)}=2$ input channels.
Conversely, the final output layer is constrained to a single channel ($N_{c,\text{out}}^{(N_l)}=1$), corresponding to the scalar concentration field.
To ensure the output respects physical constraints (\ie concentration $c \ge 0$), a ReLU activation is employed in the final layer.

The training of the CNN is formulated as an optimization problem with respect to the network parameters $\theta_C = \{\mathbf{w}, \mathbf{b}\}$.
These parameters are updated via backpropagation to minimize a cost function defined as the MSE between the estimated field and the reference LES concentration
\begin{equation}
    \theta_C^\ast = \argmin_{\theta_C} \frac{1}{N_t^\mathrm{(train)}} \sum_{k=1}^{N_t^\mathrm{(train)}} \| \mathbf{c}_\mathrm{LES}(t_k) - \mathcal{F}_\mathrm{CNN}(\mathbf{q}_\mathrm{LES}(t_k); \theta_C) \|_2^2.
    \label{eq:method:cnn:2}
\end{equation}
SGD optimization is performed utilizing the Adam algorithm to ensure efficient convergence.
By training on diverse flow snapshots, specific vortex configurations are associated with their corresponding pollutant distribution patterns by the network.

% ---------------------------------------------------------------------------
\subsection{ROM workflow}
% ---------------------------------------------------------------------------
\label{sec:method:rom}

\begin{figure}
    \centering
    \includegraphics[width=1\textwidth]{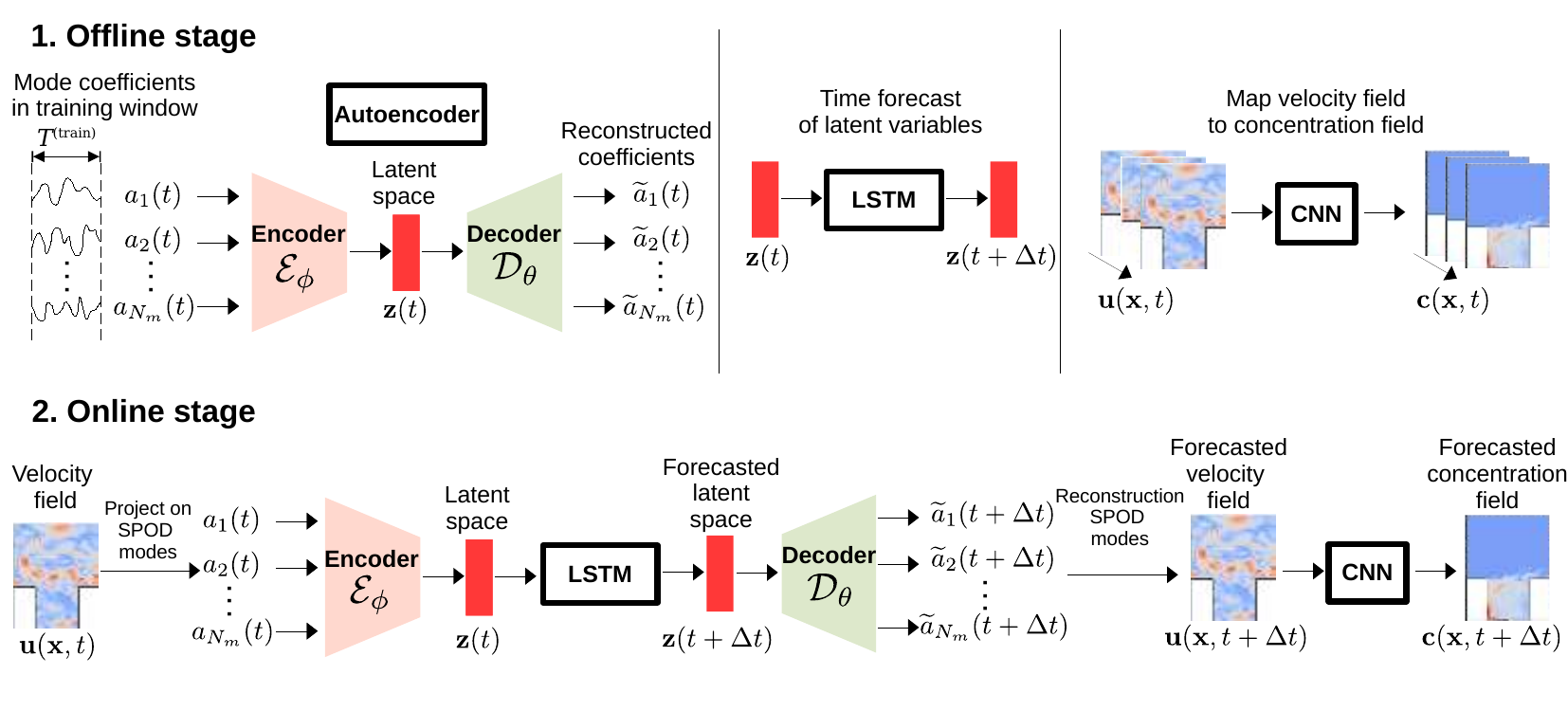}
    \caption{
    Schematic of the data-driven ROM framework.
    The offline stage involves training the three independent neural network architectures -- AE, LSTM and CNN.
    The online stage utilizes the trained models to forecast the velocity and concentration fields from an initial condition.
    }
    \label{fig:method:rom:1}
\end{figure}

The individual components detailed in Sections \ref{sec:method:spod}-\ref{sec:method:cnn} are integrated to form the complete data-driven ROM framework, as illustrated in \fref{fig:method:rom:1}.
The workflow is divided into two distinct phases: an \emph{offline} training phase and an \emph{online} prediction phase.

In the \emph{offline phase}, the high-fidelity LES dataset is utilized to construct the reduced basis and train the neural network architectures.
First, the SPOD modes and the corresponding time-domain coefficients are computed from the velocity snapshots.
These coefficients serve as the training data for the autoencoder (AE), which is optimized to produce the compressed latent space vectors.
Subsequently, the LSTM network is trained to learn the temporal dynamics of this latent space.
Independently, the convolutional neural network (CNN) is trained to map the reconstructed velocity fields onto the corresponding pollutant concentration fields.

In the \emph{online phase}, the trained framework is deployed to forecast the flow evolution autonomously.
The system is initialized with a short sequence of velocity fields (corresponding to the LSTM look-back window).
These initial fields are projected onto the SPOD basis and compressed via the AE encoder to obtain the initial latent sequence.
This sequence is then propagated forward in time using the LSTM in a recursive loop.
At each time step, the forecasted latent vector is passed through the AE decoder to recover the SPOD coefficients, which are then used to reconstruct the full velocity field.
Finally, the pollutant concentration is derived by passing the reconstructed velocity field through the trained CNN.
The performance of this framework is evaluated using the street canyon dataset described in \sref{sec:numsim}, with results discussed in the following section.

%% file: section-results.tex
% ---------------------------------------------------------------------------
\section{Results}
% ---------------------------------------------------------------------------
\label{sec:result}

% ---------------------------------------------------------------------------
\subsection{SPOD of canyon flow data}
% ---------------------------------------------------------------------------
\label{sec:result:spod}

The snapshots obtained from the street canyon LES (\sref{sec:numsim}) are decomposed to extract the SPOD modes.
The calculations are performed using the \texttt{PySPOD} library \citep{rogowskiUnlockingMassivelyParallel2024}, which provides a parallelized implementation of Welch's method (\sref{sec:method:spod}).
The dataset, comprising $N_t = 80000$ snapshots, is partitioned into training and testing subsets in a 9:1 ratio.
Consequently, the training set contains $N_t^\mathrm{(train)} = 72000$ snapshots, while the remaining $N_t^\mathrm{(test)} = 8000$ snapshots are reserved for testing.
The results presented in this subsection correspond to the SPOD analysis performed on the training subset.

% ---------------------------------------------------------------------------
\subsubsection{Selection of spectral estimation parameters}
% ---------------------------------------------------------------------------
\label{sec:result:spod:param}

\begin{figure}
    \centering
    \begin{subfigure}{0.49\textwidth}
        \centering
        \includegraphics[width=\textwidth]{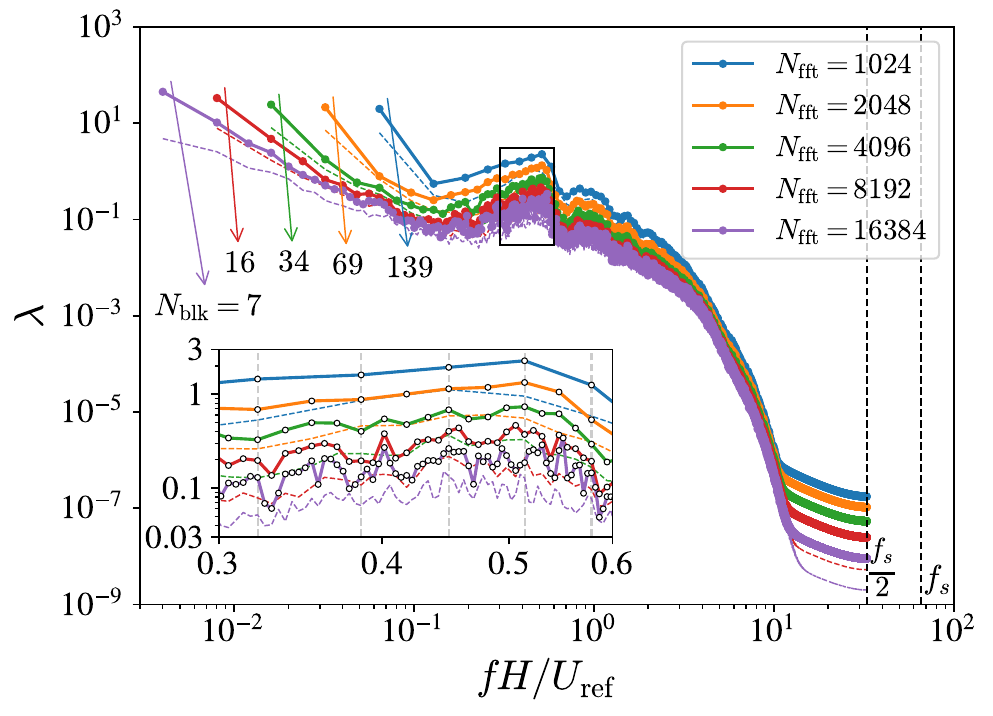}
        \caption{}
        \label{fig:result:spod:1a}
    \end{subfigure}
    \begin{subfigure}{0.49\textwidth}
        \centering
        \includegraphics[width=\textwidth]{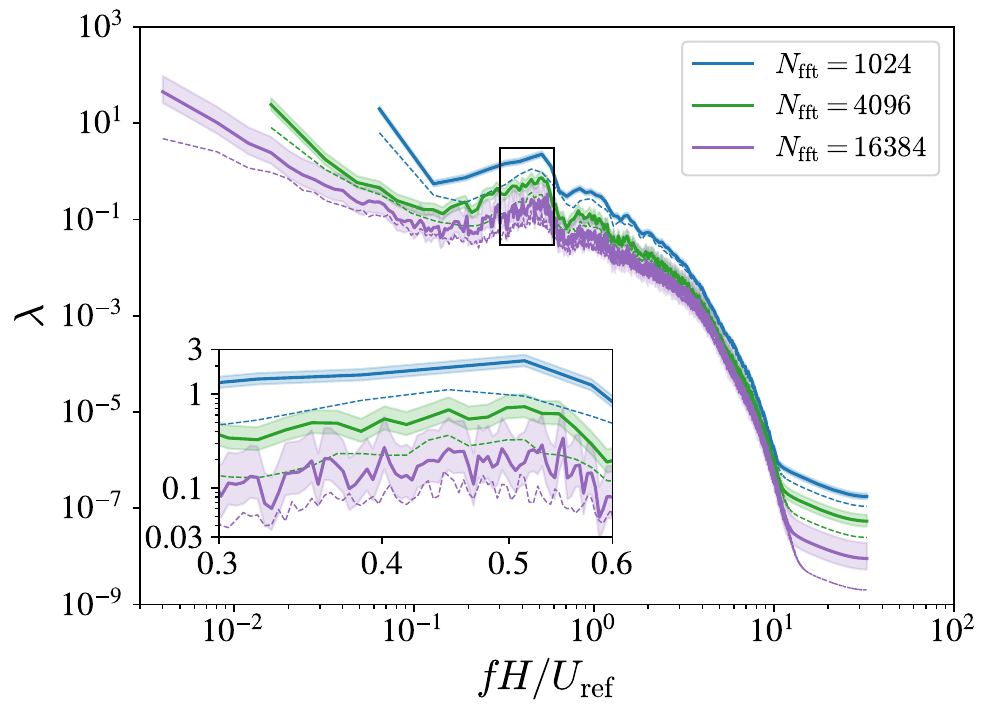}
        \caption{}
        \label{fig:result:spod:1b}
    \end{subfigure}
    \caption{
    (a) Spectrum of the two leading SPOD eigenvalues ($\lambda_{f_k}^{(1)}$ -- solid, and $\lambda_{f_k}^{(2)}$ -- dashed) for five values of the block size $N_\mathrm{fft}$.
    The numbers at the end of the colored arrows indicate the resulting number of blocks $N_\mathrm{blk}$.
    (b) Eigenvalue spectrum showing the $95\%$ confidence intervals (shaded area) for three values of $N_\mathrm{fft}$.
    }
    \label{fig:result:spod:1}
\end{figure}

The accuracy of the SPOD spectrum depends on the trade-off between frequency resolution and statistical convergence.
While a larger block size $N_\mathrm{fft}$ improves the frequency resolution, calculated as $\Delta f = f_s / N_\mathrm{fft}$, it reduces the number of available blocks $N_\mathrm{blk}$ for ensemble averaging.
The number of blocks is given by
\begin{equation}
    N_\mathrm{blk} = \flr*{\frac{N_t^\mathrm{(train)} - N_\mathrm{ovlp}}{N_\mathrm{fft} - N_\mathrm{ovlp}}},
    \label{eq:result:spod:param:1}
\end{equation}
where $\flr*{\cdot}$ is the floor operation and $N_\mathrm{ovlp}$ is the overlap between segments.
To identify the optimal parameters for the current dataset, a sensitivity analysis was conducted.
Figure \ref{fig:result:spod:1a} displays the leading eigenvalues ($\lambda_{f_k}^{(1)}$ and $\lambda_{f_k}^{(2)}$) computed for five distinct values of $N_\mathrm{fft}$.
It is observed that increasing $N_\mathrm{fft}$ improves the resolution at low frequencies ($f^\ast < 0.1$, where $f^\ast = fH/U_\mathrm{ref}$ is the reduced frequency), which is critical for capturing slow-evolving large-scale structures.
For lower values of $N_\mathrm{fft}$, the spectrum appears artificially elevated because the total energy (the area under the curve) is distributed over fewer discrete frequency bins.
However, increasing $N_\mathrm{fft}$ significantly reduces $N_\mathrm{blk}$.
As demonstrated in \fref{fig:result:spod:1b}, for high values of $N_\mathrm{fft}$ (where $N_\mathrm{blk}$ is low), the $95\%$ confidence intervals widen significantly.
This indicates that the spectral estimate is not statistically converged due to an insufficient number of realizations for averaging.

A compromise between resolution and convergence was found by selecting $N_\mathrm{fft} = 4096$ with a $50\%$ overlap ($N_\mathrm{ovlp} = 2048$).
This configuration yields $N_\mathrm{blk} = 34$ blocks, ensuring sufficiently narrow confidence intervals while providing a frequency resolution of $\Delta f \approx 0.016 U_\mathrm{ref}/H$.
This resolution is fine enough to capture the fundamental shedding frequencies of the canyon shear layer without aliasing the lowest energetic scales.
Furthermore, the Nyquist frequency of $500~\mathrm{Hz}$ ($\approx 32 U_\mathrm{ref}/H$) is sufficient to cover the relevant inertial range of the turbulence.

% ---------------------------------------------------------------------------
\subsubsection{Analysis of SPOD spectrum}
% ---------------------------------------------------------------------------
\label{sec:result:spod:spectrum}

\begin{figure}
    \centering
    \begin{subfigure}{0.4\textwidth}
        \centering
        \includegraphics[width=\textwidth]{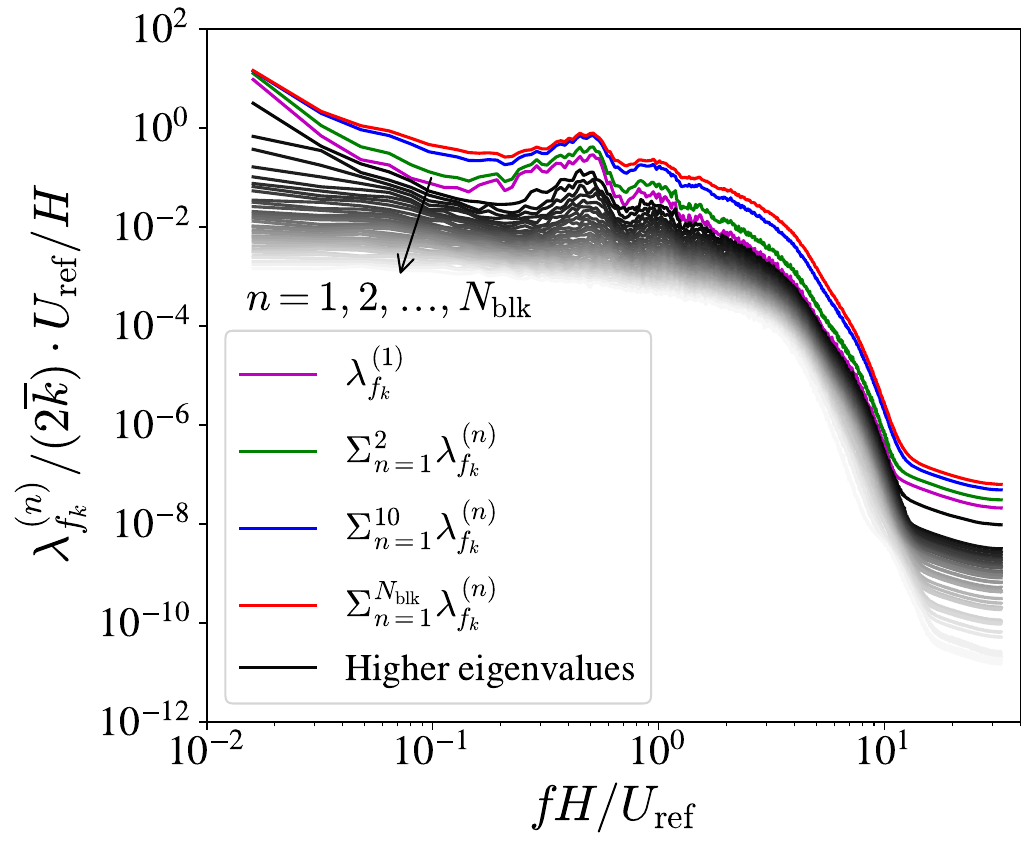}
        \caption{}
        \label{fig:result:spod:2a}
    \end{subfigure}
    \begin{subfigure}{0.52\textwidth}
        \centering
        \includegraphics[width=\textwidth]{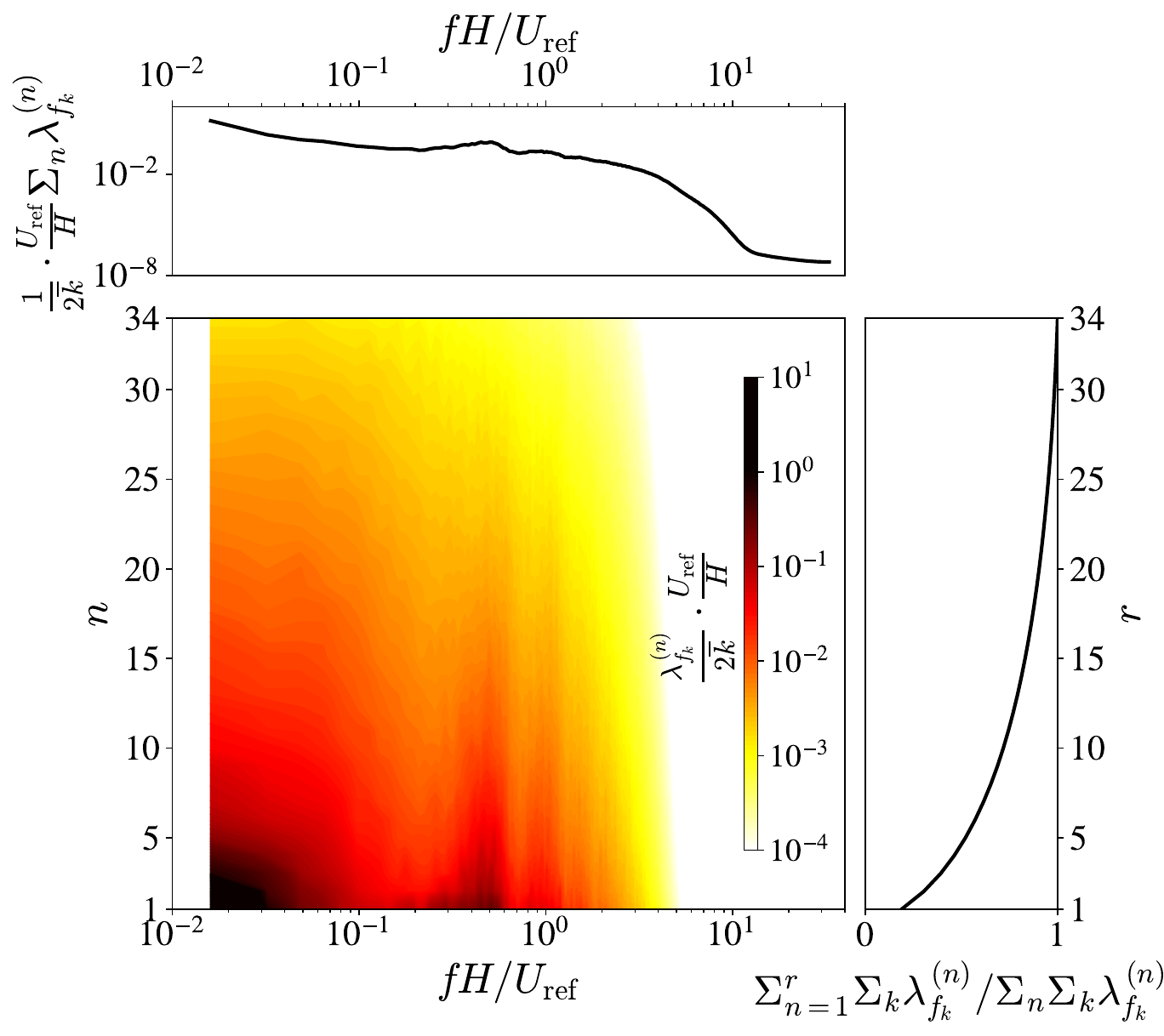}
        \caption{}
        \label{fig:result:spod:2b}
    \end{subfigure}
    \caption{
    (a) Eigenvalue spectrum of SPOD modes for $N_\mathrm{fft}=4096$ and $N_\mathrm{blk}=34$.
    (b) Contour of nondimensional eigenvalues with respect to both the frequency and the spatial mode number.
    The summed eigenvalues along the frequency and the spatial mode number are also plotted in the top and right insets respectively.
    Here $\sum_k$ and $\sum_n$ denote summation of the eigenvalues over all the discrete frequencies and modes.
    }
    \label{fig:result:spod:2}
\end{figure}

\begin{figure}
    \centering
    \begin{subfigure}{0.24\textwidth}
        \centering
        \includegraphics[width=\textwidth]{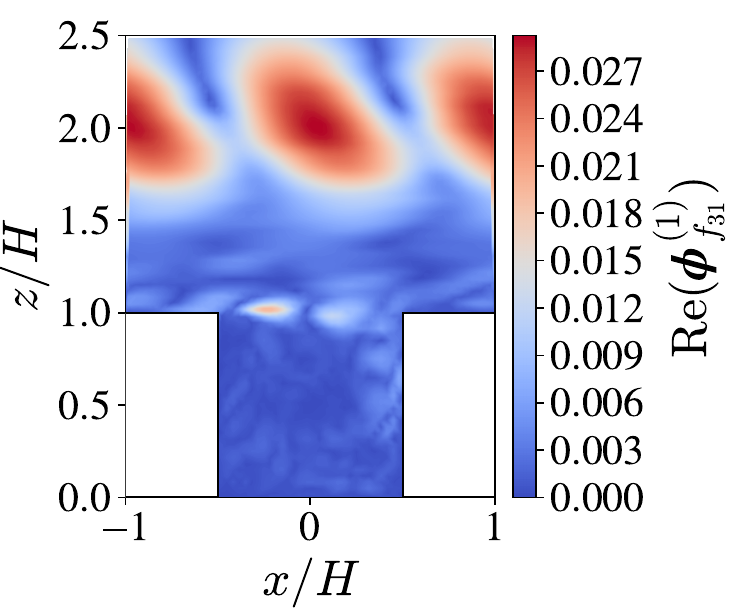}
        \caption{$f^\ast=0.5$, Mode 1}
        \label{fig:result:spod:3a}
    \end{subfigure}
    \begin{subfigure}{0.24\textwidth}
        \centering
        \includegraphics[width=\textwidth]{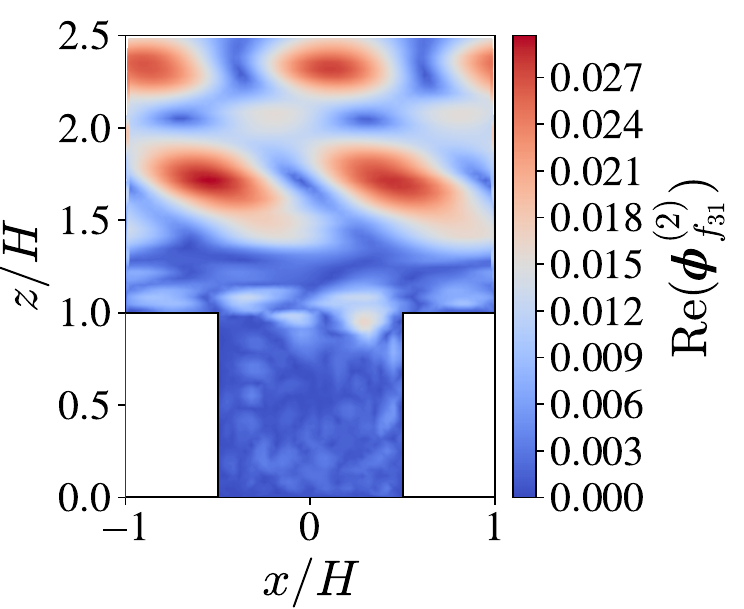}
        \caption{$f^\ast=0.5$, Mode 2}
        \label{fig:result:spod:3b}
    \end{subfigure}
    \begin{subfigure}{0.24\textwidth}
        \centering
        \includegraphics[width=\textwidth]{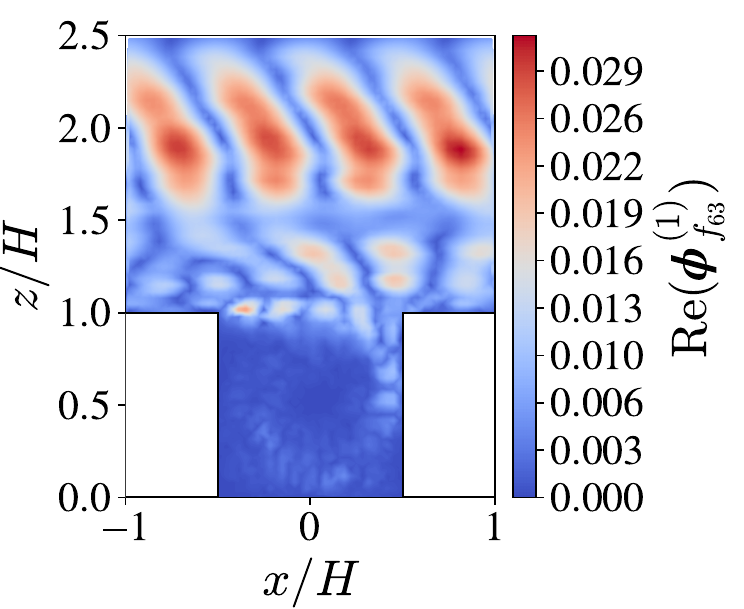}
        \caption{$f^\ast=1.0$, Mode 1}
        \label{fig:result:spod:3c}
    \end{subfigure}
    \begin{subfigure}{0.24\textwidth}
        \centering
        \includegraphics[width=\textwidth]{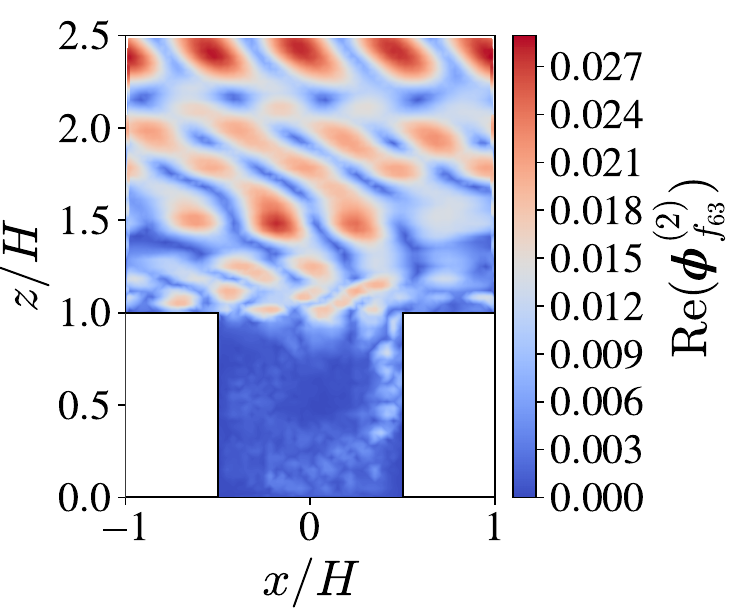}
        \caption{$f^\ast=1.0$, Mode 2}
        \label{fig:result:spod:3d}
    \end{subfigure}
    \caption{
    Real component of the SPOD modes at reduced frequencies (a,b) $f^\ast = 0.5$ and (c,d) $f^\ast=1.0$.
    The leading optimal modes $\boldsymbol{\phi}_{f_k}^{(1)}$ are shown in (a,c), while the suboptimal modes $\boldsymbol{\phi}_{f_k}^{(2)}$ are shown in (b,d).
    }
    \label{fig:result:spod:3}
\end{figure}

The SPOD eigenvalue spectrum of the canyon flow data is presented in \fref{fig:result:spod:2a}.
The total energy corresponds to the sum of the eigenvalues across all frequencies and modes.
The dominant eigenvalue $\lambda_{f_k}^{(1)}$ (magenta line) represents the energy contained in the optimal mode at each frequency.
It is observed that the energy is rapidly decaying with mode rank $n$; the gray lines representing higher-order modes ($n>1$) show significantly lower energy content.
This spectral hierarchy is further quantified in \fref{fig:result:spod:2b}, which maps the energy distribution across frequency and mode number.
It is confirmed that the entirety of the turbulent kinetic energy (TKE) is captured by the $N_\mathrm{blk}$ modes.
Notably, the energy is concentrated in the low-frequency range ($f^\ast < 1.0$) and low mode numbers ($n < 5$), suggesting that the essential flow dynamics can be reconstructed using a severely truncated basis.

Two distinct spectral peaks are identified in the leading modes at frequencies $f^\ast = 0.5$ and $f^\ast = 1.0$.
The physical structures associated with these peaks are visualized in \fref{fig:result:spod:3}, where the real component of the complex spatial modes is plotted.
At $f^\ast = 0.5$, the optimal mode (Mode 1, \fref{fig:result:spod:3a}) reveals a large-scale structure filling the entire streamwise extent of the domain above the canyon.
Given the domain length $L_x = 2H$ and reference velocity $U_\mathrm{ref}$, this frequency corresponds to the convective timescale of the simulation box ($f = U_\mathrm{ref}/L_x \implies f^\ast = 0.5$).
Consequently, this mode is interpreted as a domain-scale periodicity artifact (\dquotes{box mode}) inherent to the streamwise periodic boundary conditions, rather than a physical phenomenon.
However, this mode also captures the low-frequency pumping interaction between the external flow and the shear layer at the roof level.

At $f^\ast = 1.0$, the optimal mode (\fref{fig:result:spod:3c}) exhibits a coherent train of vortices originating at the leading edge of the canyon ($z/H=1$).
This frequency corresponds to the Kelvin-Helmholtz (KH) instability mechanism in the shear layer \citep{cuiLargeeddySimulationTurbulent2004}.
These vortices are the primary drivers of momentum and scalar exchange between the canyon cavity and the external flow.
The suboptimal modes ($n=2$) at these frequencies (\fref{fig:result:spod:3b} and \ref{fig:result:spod:3d}) display more complex, multilobed structures.
These higher-order modes generally represent statistical corrections required to reconstruct the stochasticity of the turbulence, rather than distinct physical instabilities.
It is noted that for reduced-order modeling, the exact physical interpretation of these higher modes is secondary to their ability to efficiently capture the variance of the flow field.

The full SPOD feature space comprises $N_\mathrm{fc} \times N_\mathrm{blk} \approx 7 \times 10^4$ degrees of freedom.
This high dimensionality necessitates the application of pruning techniques before the ROM can be efficiently trained.
The dimensionality reduction process is detailed in the following section.

% ---------------------------------------------------------------------------
\subsubsection{Dimensionality reduction of the SPOD space}
% ---------------------------------------------------------------------------
\label{sec:result:spod:dim}

\begin{figure}
    \centering
    \begin{subfigure}{0.35\textwidth}
        \centering
        \includegraphics[width=\textwidth]{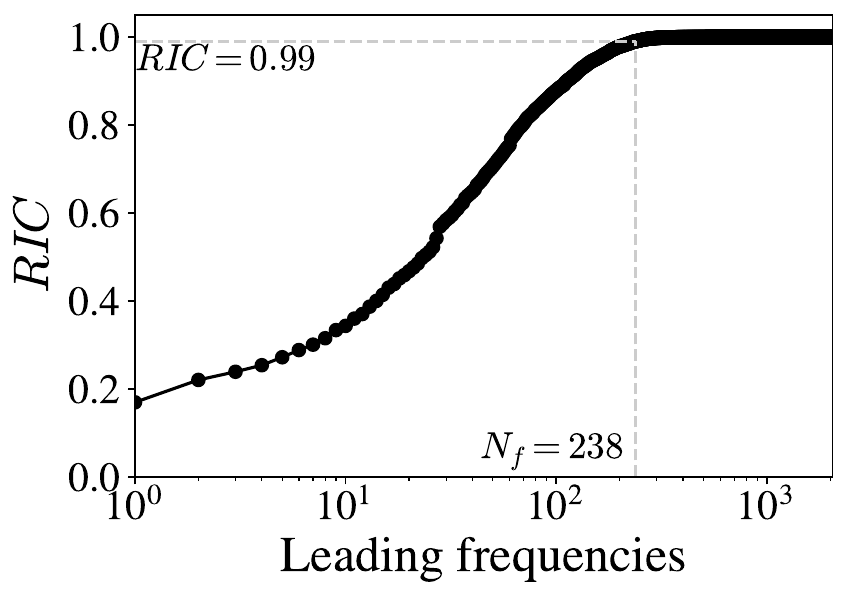}
        \caption{}
        \label{fig:result:spod:4a}
    \end{subfigure}
    \hspace{2em}
    \begin{subfigure}{0.35\textwidth}
        \centering
        \includegraphics[width=\textwidth]{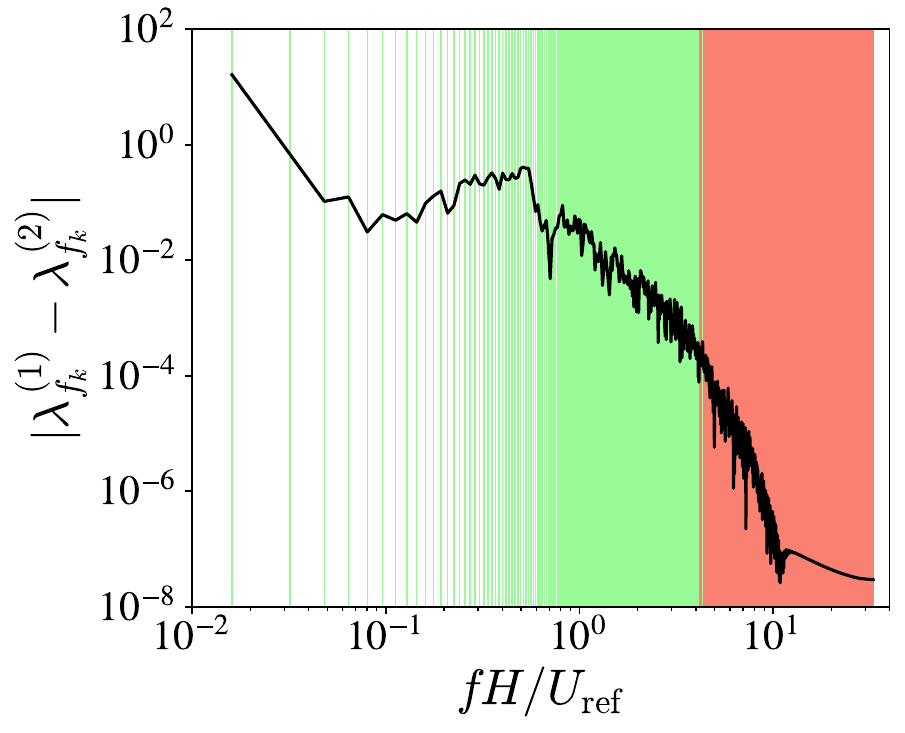}
        \caption{}
        \label{fig:result:spod:4b}
    \end{subfigure}
    \caption{
    (a) Relative information content (RIC) as a function of the number of retained leading frequencies.
    (b) Eigenvalue separation $\Delta \lambda_{f_k}$ spectrum.
    The green and red vertical lines indicate the preserved and discarded frequencies respectively.
    }
    \label{fig:result:spod:4}
\end{figure}

\begin{figure}
    \centering
    \begin{subfigure}{0.38\textwidth}
        \centering
        \includegraphics[width=\textwidth]{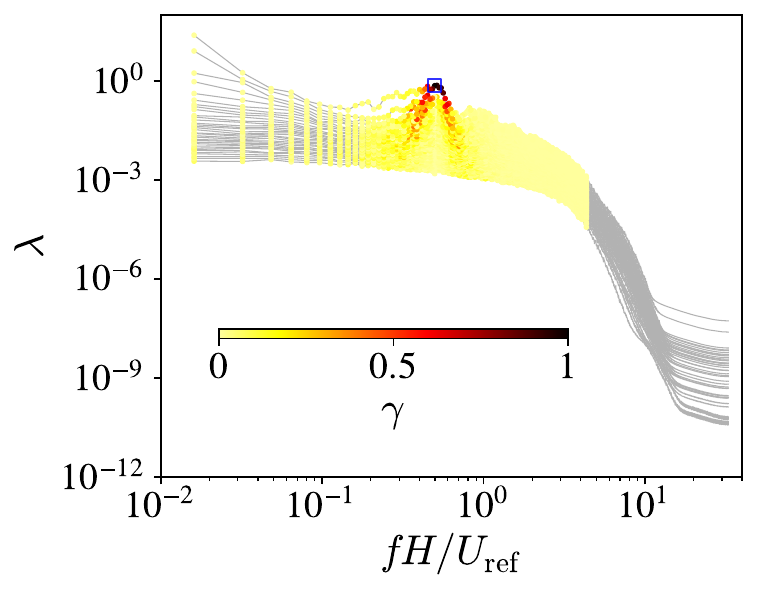}
        \caption{}
        \label{fig:result:spod:5a}
    \end{subfigure}
    \begin{subfigure}{0.3\textwidth}
        \centering
        \includegraphics[width=\textwidth]{modes_U_all_St0031_mode0001_crop_comp}
        \caption{$\gamma = 1.00$ (Target)} % $\gamma_{(31,1), (31,1)} = 1.00$
        \label{fig:result:spod:5b}
    \end{subfigure}
    \begin{subfigure}{0.3\textwidth}
        \centering
        \includegraphics[width=\textwidth]{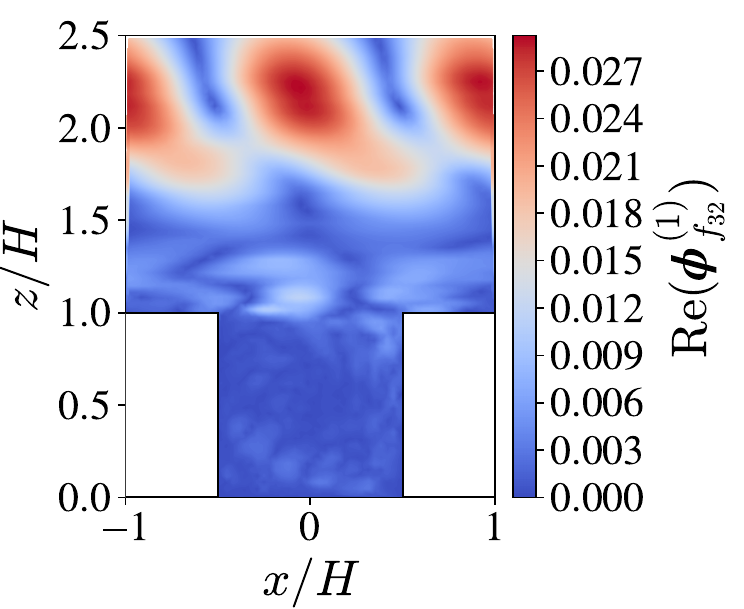}
        \caption{$\gamma = 0.94$ (Neighbor)} % $\gamma_{(31,1), (32,1)} = 0.94$
        \label{fig:result:spod:5c}
    \end{subfigure}
    \caption{
    (a) Spatial similarity coefficient $\gamma$ computed with respect to the target mode $\boldsymbol{\phi}_{f_{31}}^{(1)}$ (indicated by the blue square).
    (b) Real part of the target mode.
    (c) Real part of the most similar neighboring mode ($\boldsymbol{\phi}_{f_{32}}^{(1)}$), illustrating the redundancy in spatial structures across frequencies.
    }
    \label{fig:result:spod:5}
\end{figure}

To construct a tractable reduced-order model, the dimension of the SPOD feature space must be reduced.
This was achieved by sequentially applying the two criteria defined in \sref{sec:method:spod:dim}.

First, the spectral energy criterion was applied to truncate the frequency space.
The eigenvalues were sorted by the separation magnitude $\Delta \lambda_{f_k}$, and the relative information content (RIC) was computed.
As shown in \fref{fig:result:spod:4a}, a threshold of $\varepsilon_\text{RIC} = 0.99$ was selected.
This resulted in the retention of $N_f = 238$ leading frequencies.
The distribution of these preserved frequencies is visualized in \fref{fig:result:spod:4b} (green zone).
It is observed that the selection effectively isolates the energetically significant low-frequency band, while the high-frequency tail (red zone), which contributes negligibly to the total variance, is discarded.
This step reduced the feature space to $N_f \times N_\mathrm{blk} \approx 8000$ modes, representing a reduction of approximately $88\%$.

Second, the spatial similarity criterion was applied to remove redundant modes.
The necessity of this step is illustrated in \fref{fig:result:spod:5}.
The similarity map for a target mode at $f^\ast=0.5$ (\fref{fig:result:spod:5a}) reveals strong correlations ($\gamma > 0.9$) with modes at adjacent frequencies.
Visually, the target mode (\fref{fig:result:spod:5b}) and its neighbor (\fref{fig:result:spod:5c}) exhibit nearly identical shear layer topologies.
This implies that multiple modes are encoding the same physical structure as it evolves slowly through the frequency domain.

\begin{figure}
    \centering
    \begin{subfigure}{0.32\textwidth}
        \centering
        \includegraphics[width=0.96\textwidth]{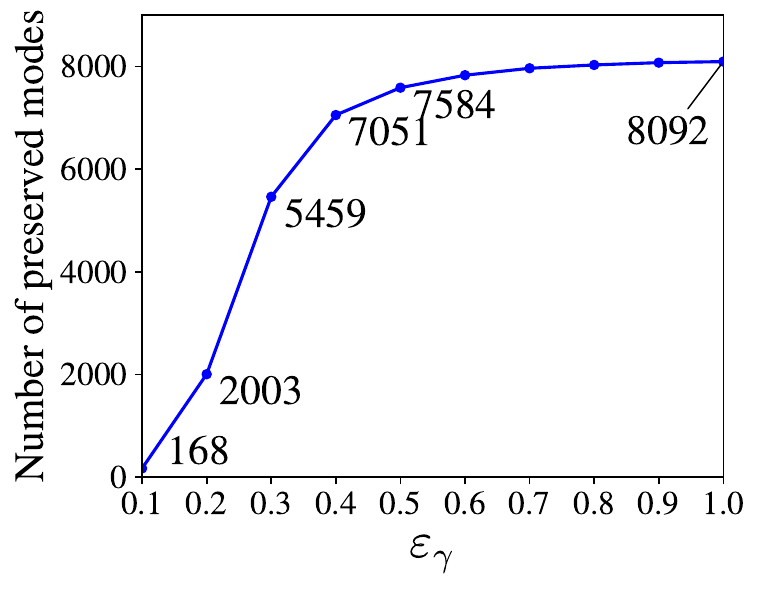}
        \caption{}
        \label{fig:result:spod:6a}
    \end{subfigure}
    \begin{subfigure}{0.32\textwidth}
        \centering
        \includegraphics[width=0.97\textwidth]{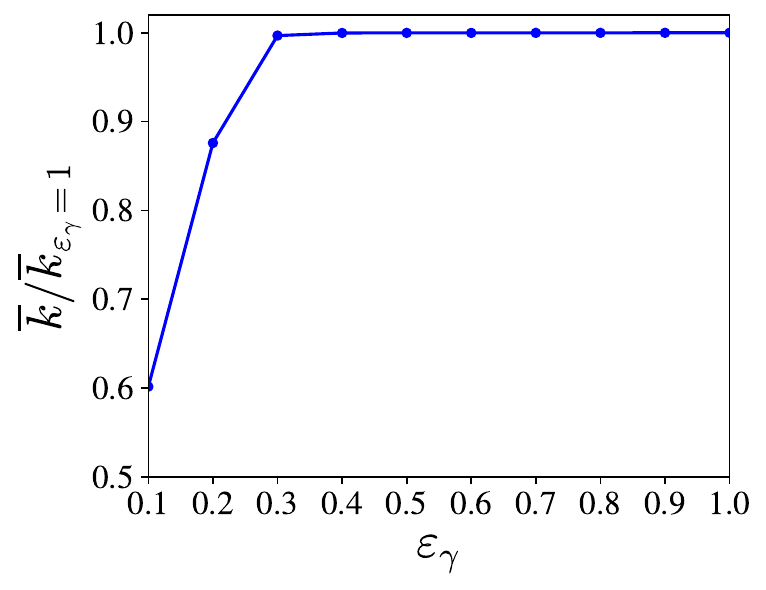}
        \caption{}
        \label{fig:result:spod:6b}
    \end{subfigure}
    \begin{subfigure}{0.32\textwidth}
        \centering
        \includegraphics[width=\textwidth]{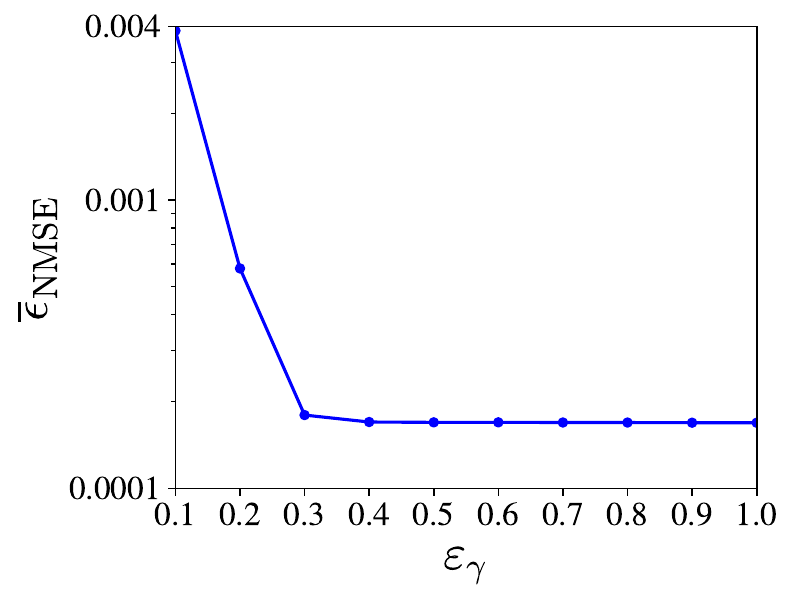}
        \caption{}
        \label{fig:result:spod:6c}
    \end{subfigure}
    \caption{
    Sensitivity analysis for the similarity threshold $\varepsilon_\gamma$: (a) Number of preserved modes; (b) Fraction of preserved TKE relative to the full basis; (c) Normalized mean-squared error (NMSE) of the reconstruction.
    }
    \label{fig:result:spod:6}
\end{figure}

To remove this redundancy, a sensitivity analysis on the similarity threshold $\varepsilon_\gamma$ was conducted (\fref{fig:result:spod:6}).
As $\varepsilon_\gamma$ is lowered, the number of retained modes decreases non-linearly (\fref{fig:result:spod:6a}), while the preserved TKE remains practically constant for $\varepsilon_\gamma \ge 0.3$ (\fref{fig:result:spod:6b}), tending toward the total kinetic energy.
A threshold of $\varepsilon_\gamma = 0.2$ was selected.
Although this aggressive pruning increases the reconstruction error (\fref{fig:result:spod:6c}), it reduces the final basis size to $N_m = 2003$ modes, \ie a $97\%$ reduction from the original space.

\begin{figure}
    \centering
    \begin{minipage}{0.49\textwidth}
        \centering
        \begin{subfigure}{0.55\linewidth}
            \centering
            \includegraphics[width=\textwidth]{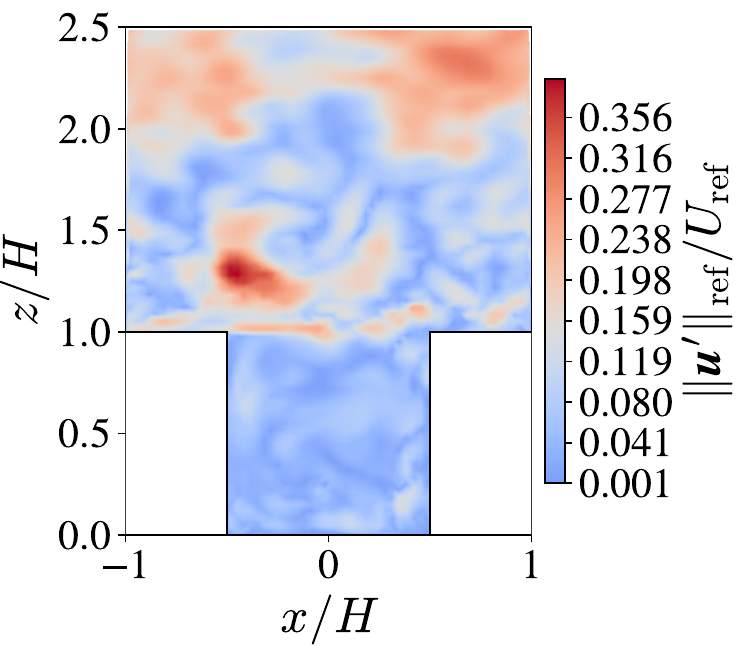}
            \caption{Reference CFD}
            \label{fig:result:spod:8a}
        \end{subfigure}
    \end{minipage}
    \begin{minipage}{0.49\textwidth}
        \begin{subfigure}{\linewidth}
            \centering
            \includegraphics[width=\textwidth]{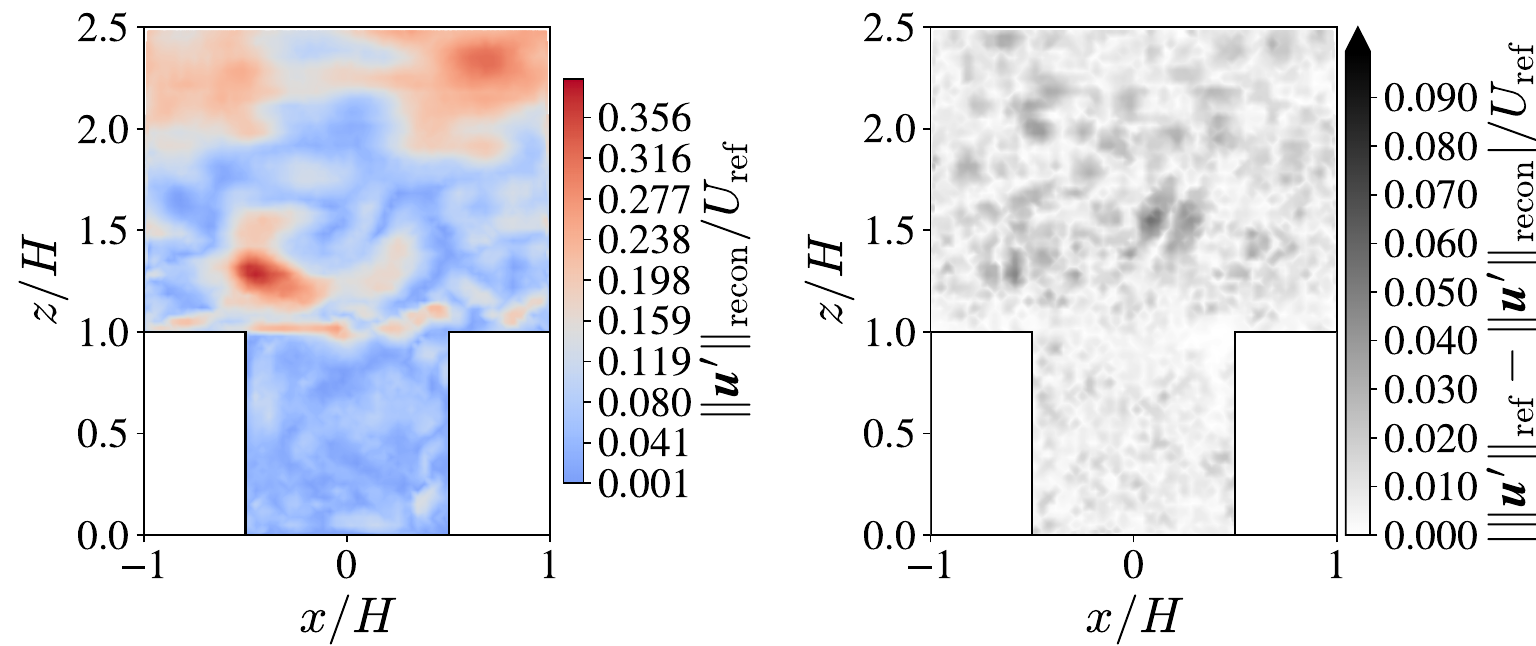}
            \caption{Reconstruction, $\varepsilon_\gamma = 1.0$}
            \label{fig:result:spod:8b}
        \end{subfigure}\\[1ex]
    \end{minipage}
    \begin{minipage}{0.49\textwidth}
        \begin{subfigure}{\linewidth}
            \centering
            \includegraphics[width=\textwidth]{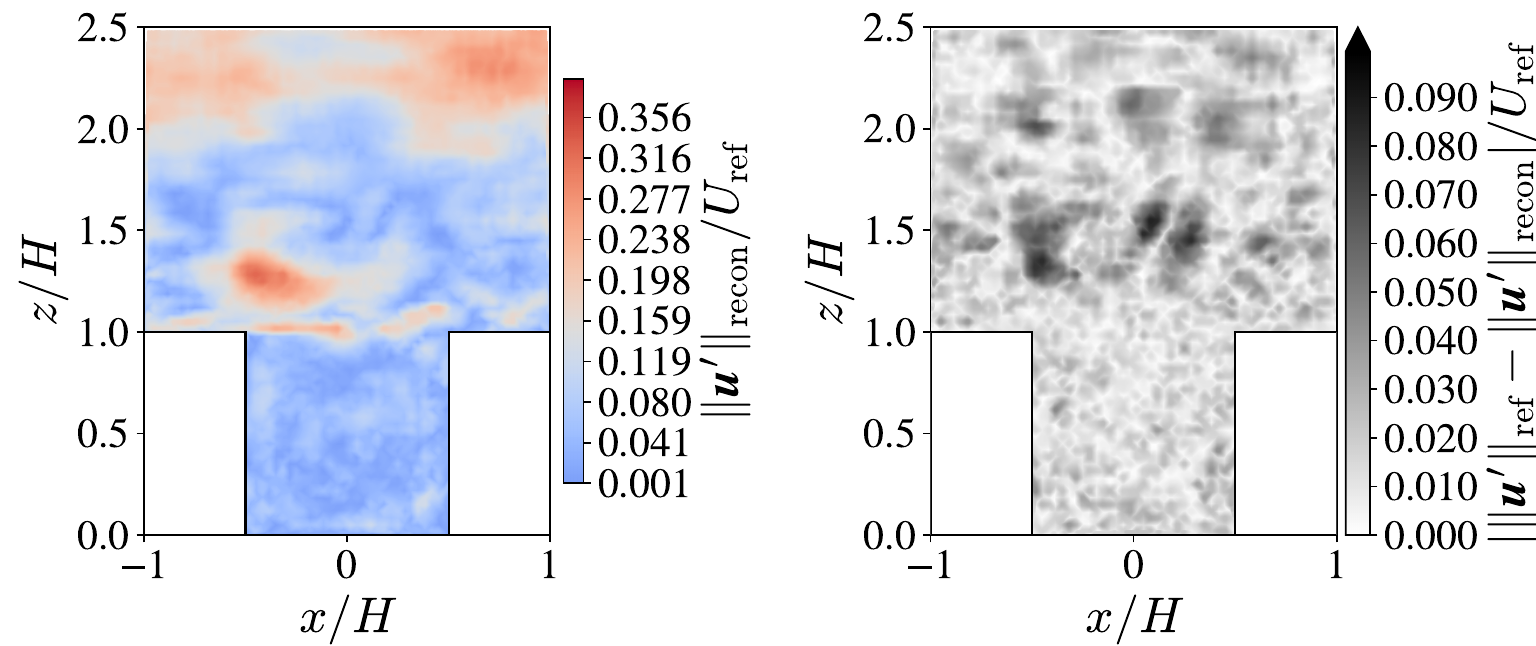}
            \caption{Reconstruction, $\varepsilon_\gamma = 0.2$}
            \label{fig:result:spod:8c}
        \end{subfigure}
    \end{minipage}
    \begin{minipage}{0.49\textwidth}
        \begin{subfigure}{\linewidth}
            \centering
            \includegraphics[width=\textwidth]{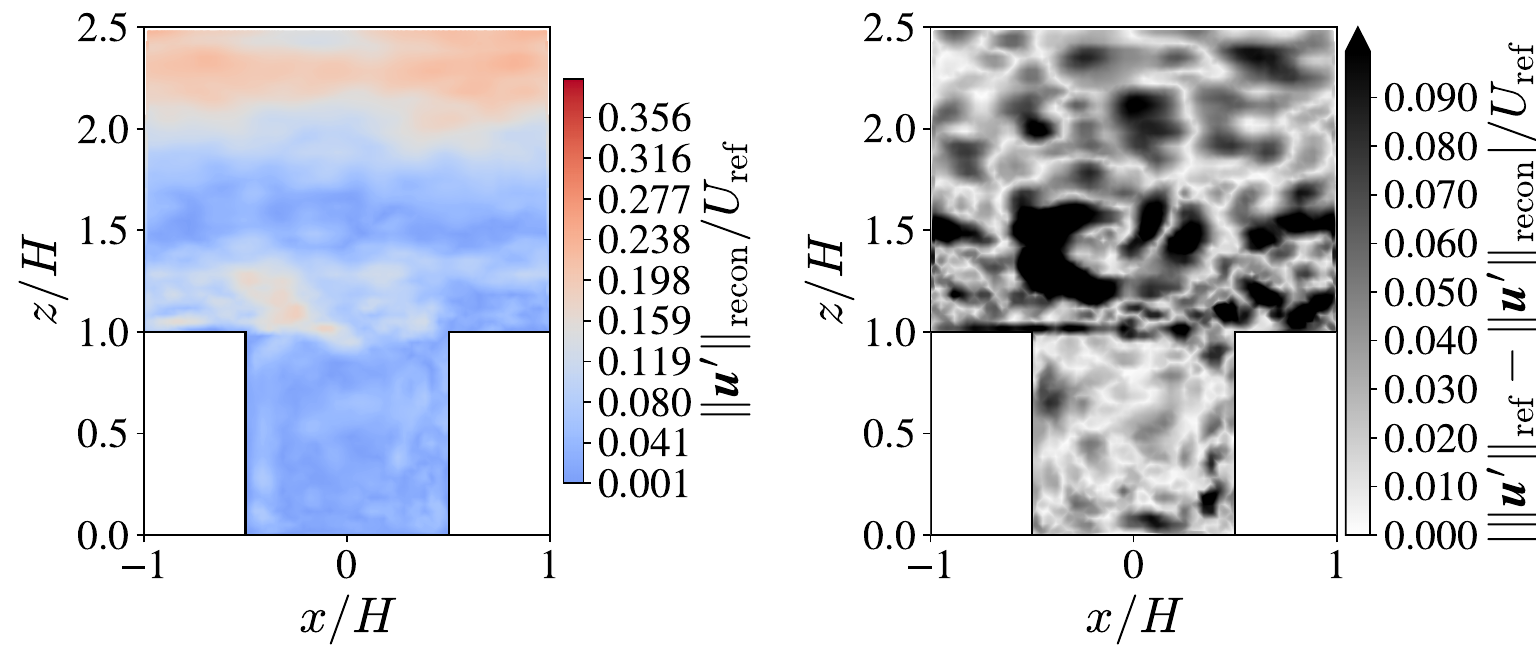}
            \caption{Reconstruction, $\varepsilon_\gamma = 0.1$}
            \label{fig:result:spod:8d}
        \end{subfigure}
    \end{minipage}
    \caption{
    Instantaneous velocity fluctuation fields at $tU_\mathrm{ref}/H = 540$.
    (a) Reference field. (b-d) Reconstructions using bases pruned with $\varepsilon_\gamma = 1.0, 0.2, 0.1$.
    The right column in each panel displays the local $L_1$-error.
    }
    \label{fig:result:spod:8}
\end{figure}

\begin{figure}
    \centering
    \includegraphics[width=.5\textwidth]{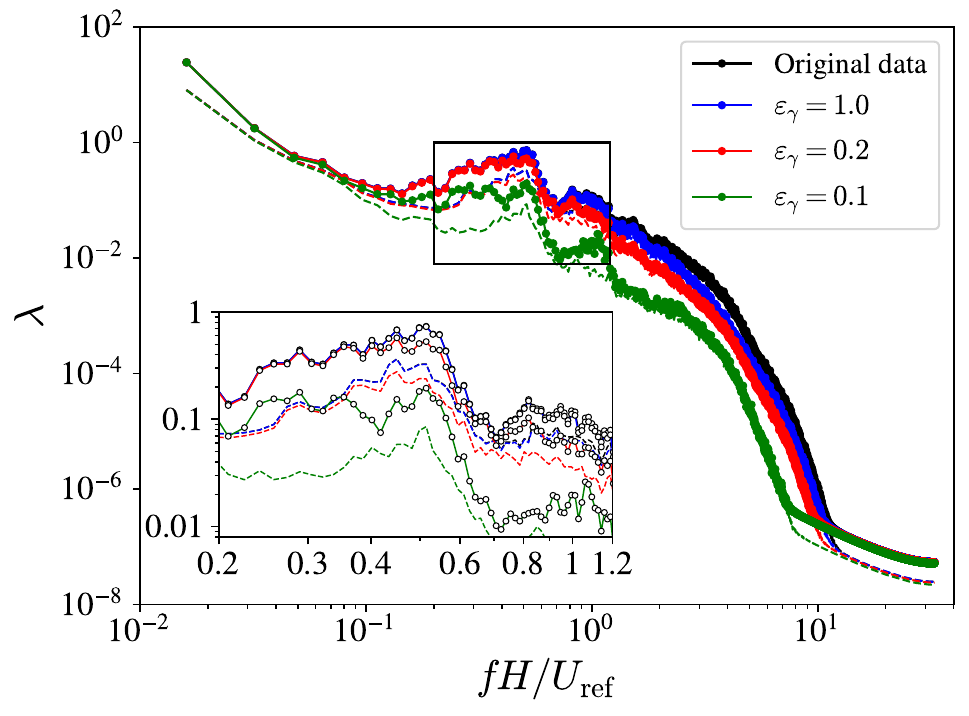}
    \caption{
    Comparison of the SPOD eigenvalue spectra for different levels of reconstruction of velocity fields ($\varepsilon_\gamma=1.0, 0.2, 0.1$) to reference data (LES). The solid lines correspond to the first mode and the dashed lines to the second mode. 
    }
    \label{fig:result:spod:7}
\end{figure}

The fidelity of this reduced basis was verified in both physical and  spectral space.
To this end, the different velocity components were first reconstructed, retaining only the modes corresponding to a given value of $\varepsilon_\gamma$. In physical space (\fref{fig:result:spod:8}), the reconstruction with $\varepsilon_\gamma = 0.2$ (\fref{fig:result:spod:8c}) successfully preserves the large-scale coherent structures, such as the Kelvin-Helmholtz vortices and the recirculation region, despite the loss of small-scale stochastic fluctuations.
In contrast, a threshold of $\varepsilon_\gamma = 0.1$ (\fref{fig:result:spod:8d}) leads to excessive filtering, resulting in the degradation of important structural features.
SPOD is then applied to the velocity fields obtained for different levels of approximation. The eigenvalue spectra are displayed for comparison in \fref{fig:result:spod:7}. The spectrum obtained with $\varepsilon_\gamma = 0.2$ closely follows the original spectrum at low frequencies, deviating only in the high-frequency range where small-scale energy is filtered out.
Consequently, the basis with $\varepsilon_\gamma = 0.2$ ($N_m = 2003$) was adopted for the subsequent reduced-order modeling.

The time-domain coefficients $\mathbf{A} \inC{N_m \times N_t^\mathrm{(train)}}$ were computed by projecting the training snapshots onto this pruned basis.
These coefficients serve as the training data for the autoencoder, as discussed in the next section.

% ---------------------------------------------------------------------------
\subsection{Latent space compression}
% ---------------------------------------------------------------------------
\label{sec:result:latent}

\begin{table}[t]
    \centering
    \caption{Hyperparameters for the autoencoder (AE) training.}
    \label{tab:result:latent:1}
    \begin{tabular}{ccccccc}
        \hline
        Encoder layers & Decoder layers & Activation & Batch size & Learning rate & Epochs \\ \hline
        [1024, 512, 256, 128, 64, 32] & [32, 64, 128, 256, 512, 1024] & $\tanh$ & 2048 & 0.001 & 5000 \\ \hline
    \end{tabular}
\end{table}

\begin{figure}
    \centering
    \begin{subfigure}{0.49\textwidth}
        \centering
        \includegraphics[width=\textwidth]{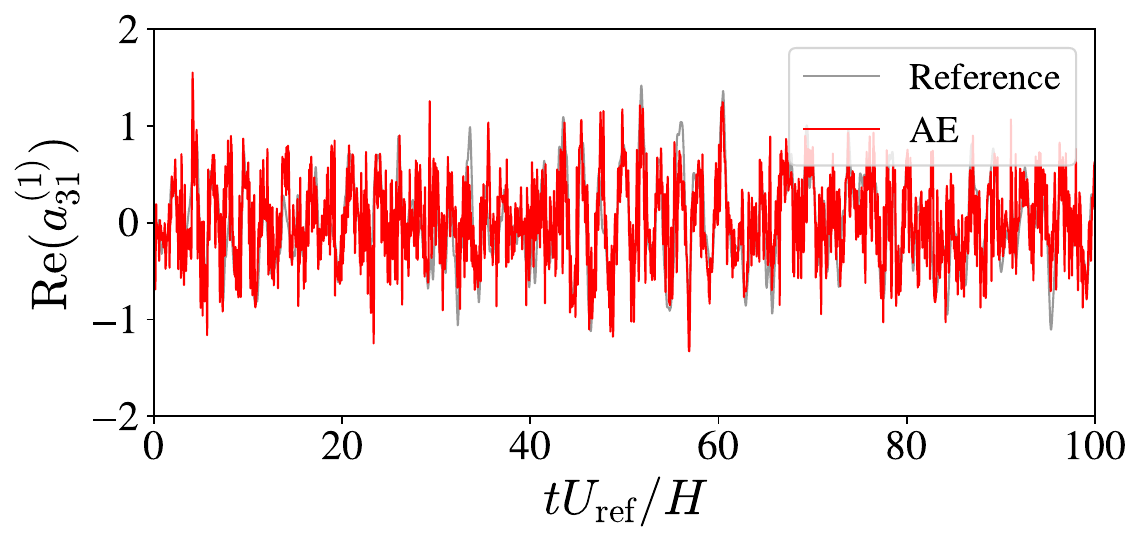}
        \caption{$N_z=5$}
        \label{fig:result:latent:1a}
    \end{subfigure}
    \begin{subfigure}{0.49\textwidth}
        \centering
        \includegraphics[width=\textwidth]{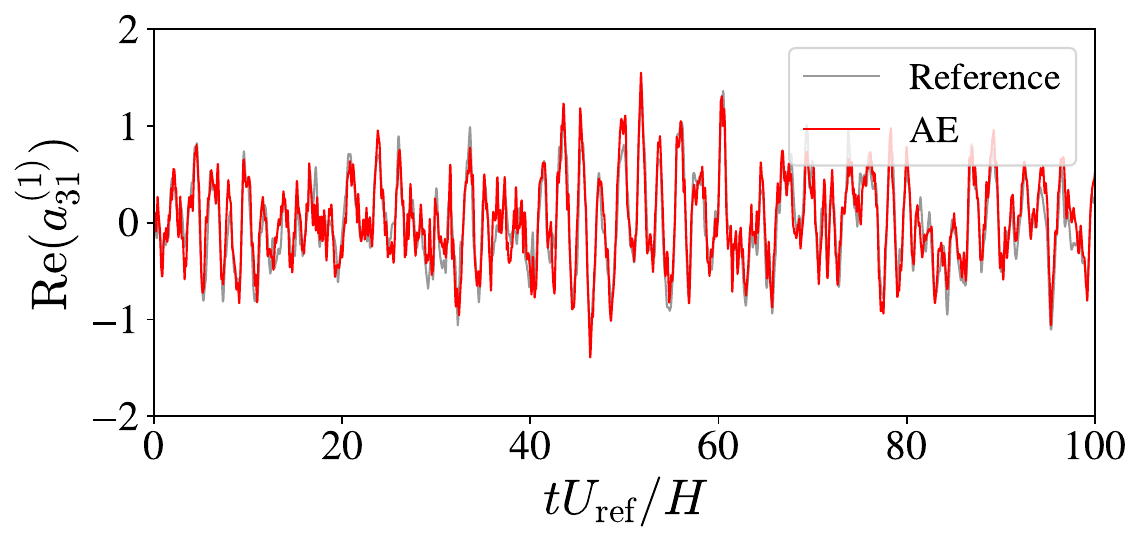}
        \caption{$N_z=30$}
        \label{fig:result:latent:1b}
    \end{subfigure}
    \caption{
    Comparison of the reconstructed time trajectory for the real component of the dominant SPOD coefficient $a_{31}^{(1)}(t)$
    (corresponding to the target mode in \fref{fig:result:spod:5a}).
    The reference SPOD trajectory is compared with AE reconstructions using latent dimensions of (a) $N_z=5$ and (b) $N_z=30$.
    }
    \label{fig:result:latent:1}
\end{figure}

\begin{figure}
    \centering
    \begin{subfigure}{0.35\textwidth}
        \centering
        \includegraphics[width=\textwidth]{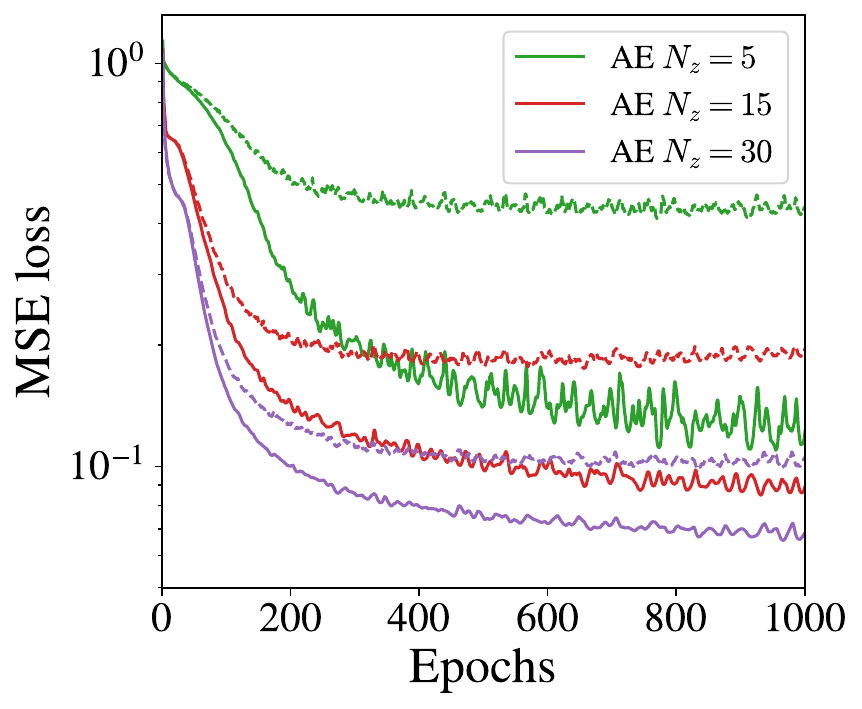}
        \caption{}
        \label{fig:result:latent:2a}
    \end{subfigure}
    \begin{subfigure}{0.6\textwidth}
        \centering
        \includegraphics[width=\textwidth]{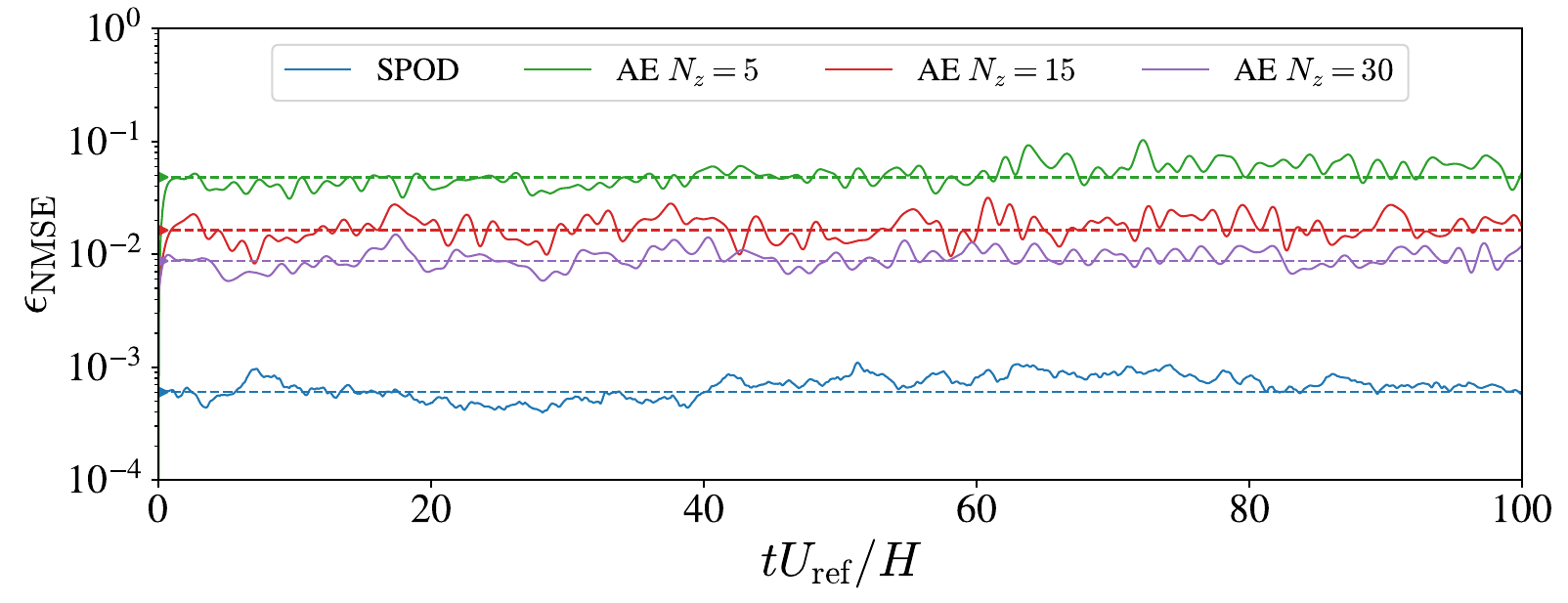}
        \caption{}
        \label{fig:result:latent:2b}
    \end{subfigure}
    \caption{
    (a) Evolution of training (solid) and validation (dashed) losses for AEs with latent sizes $N_z = 5, 15, 30$ (first 1000 epochs shown).
    (b) Time evolution of the normalized mean-squared error (NMSE) for the velocity magnitude field.
    The AE reconstruction errors are compared against the baseline SPOD projection error (blue line).
    The horizontal lines indicate the time-averaged error.
    }
    \label{fig:result:latent:2}
\end{figure}

\begin{figure}
    \centering
    \includegraphics[width=.8\textwidth]{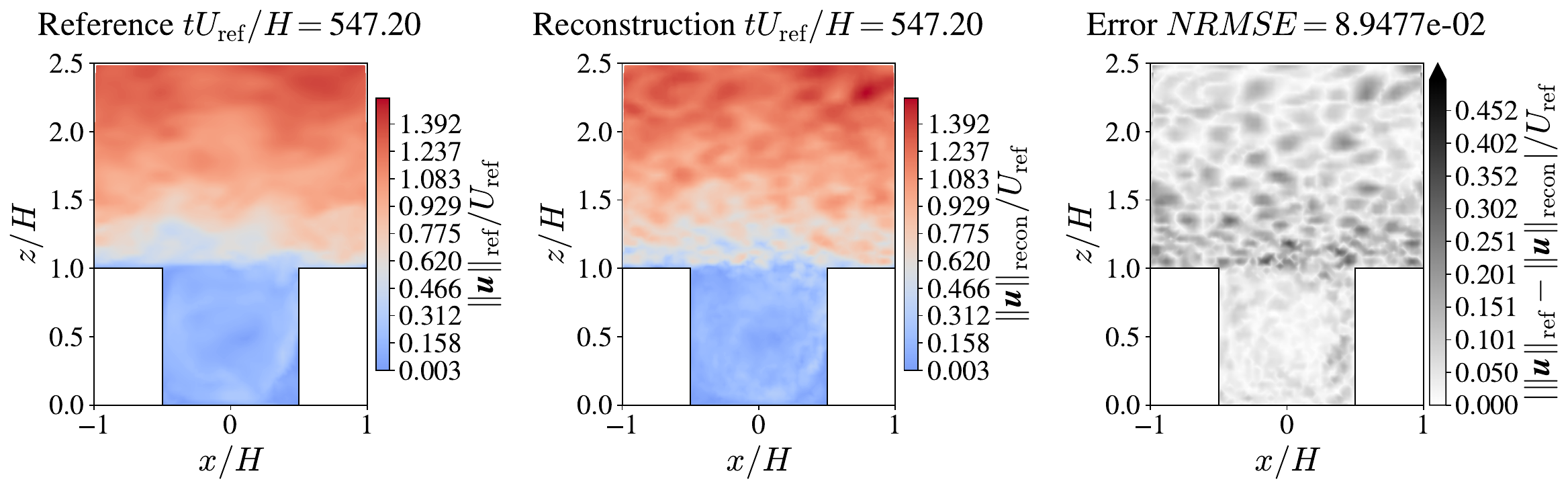}
    \caption{
    Instantaneous velocity magnitude field reconstructed using the AE with latent dimension $N_z=30$.
    }
    \label{fig:result:latent:3}
\end{figure}

As the first stage of offline modeling, the high-dimensional vector of SPOD coefficients $\mathbf{a}(t_k) \inC{N_m}$ (where $N_m = 2003$) is compressed into a compact latent vector $\mathbf{z}(t_k) \inC{N_z}$, such that $N_z \ll N_m$.
This is achieved by training the autoencoder (AE) architecture described in \sref{sec:method:ae}.
Distinct models are trained for the real and imaginary components of the coefficients.
The training dataset ($N_t^\mathrm{(train)}$) is partitioned, with $80\%$ allocated for gradient descent optimization and $20\%$ for validation to monitor generalization performance.
The specific hyperparameters used for the training are detailed in \tref{tab:result:latent:1}.

To determine the optimal compression ratio, a sensitivity study is conducted with latent dimensions of $N_z = 5, 15$, and $30$.
The impact of $N_z$ on the dynamics is illustrated in \fref{fig:result:latent:1}, which compares the reconstructed trajectory of a dominant SPOD coefficient against the ground truth.
For $N_z=5$ (\fref{fig:result:latent:1a}), the reconstructed signal does not capture the high-frequency fluctuations of the turbulent coefficients, indicating that the bottleneck is too restrictive.
Conversely, for $N_z=30$ (\fref{fig:result:latent:1b}), the reconstruction accurately tracks the reference trajectory, preserving both the phase and amplitude of the fluctuations.

This observation is quantified by the loss histories presented in \fref{fig:result:latent:2a}.
As the latent dimension is reduced, both the training and validation losses increase, reflecting the loss of information.
The consequence for the full flow field reconstruction is shown in \fref{fig:result:latent:2b}.
The AE reconstruction error is approximately an order of magnitude higher than the baseline SPOD projection error.
This increase is expected and represents the trade-off associated with nonlinear compression: the AE effectively acts as a low-pass filter, removing high-frequency content, while retaining the essential manifold of the dynamics.
Among the tested cases, $N_z=30$ yields the lowest error while maintaining a stable validation loss, indicating no overfitting.

The instantaneous velocity field reconstructed with $N_z=30$ is visualized in \fref{fig:result:latent:3}.
While some small-scale pixel-wise noise is introduced by the decoding process, the large-scale coherent structures are well-preserved, and the normalized root-mean-square error (NRMSE) remains low at $0.089$.
Consequently, a latent dimension of $N_z=30$ is selected for the subsequent temporal forecasting.

% ---------------------------------------------------------------------------
\subsection{Prediction of velocity field}
% ---------------------------------------------------------------------------
\label{sec:result:pred_u}

\begin{table}[t]
    \centering
    \caption{Search space for LSTM hyperparameter optimization.}
    \label{tab:result:pred_u:1}
    \begin{tabular}{ccccc}
        \hline
        $N_h$ & Batch size & Learning rate & Epochs & Trials \\ \hline
        [50, 200] & [500, 1000] & [0.0005, 0.0010] & 100 & 20 \\ \hline
    \end{tabular}
\end{table}

\begin{figure}
    \centering
    \begin{subfigure}{0.35\textwidth}
        \centering
        \includegraphics[width=\textwidth]{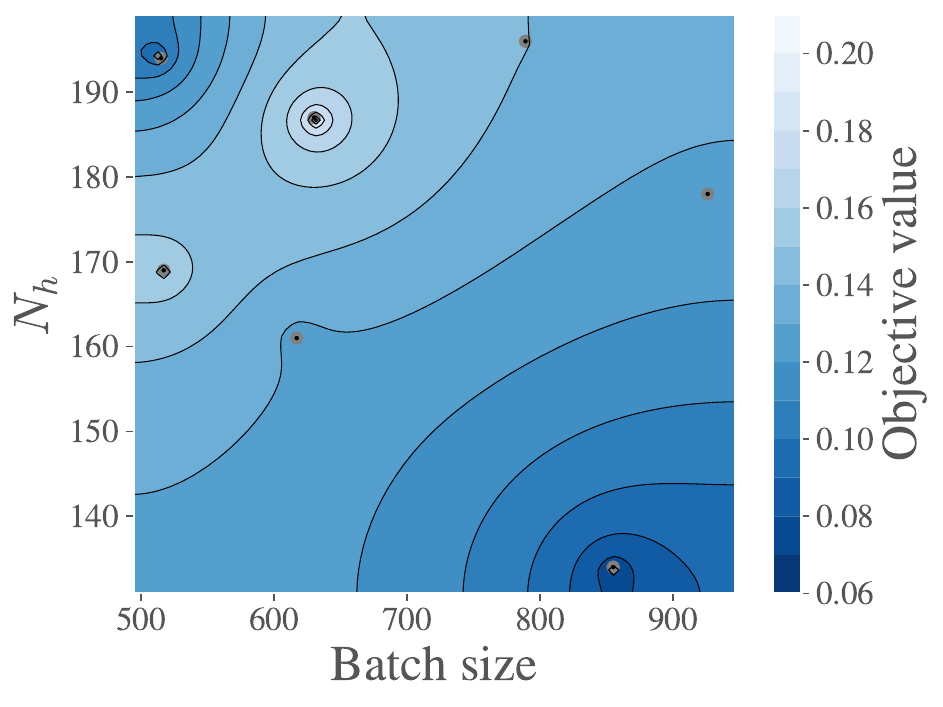}
        \caption{Real component}
        \label{fig:result:pred_u:1-0a}
    \end{subfigure}
    \hspace{3em}
    \begin{subfigure}{0.35\textwidth}
        \centering
        \includegraphics[width=\textwidth]{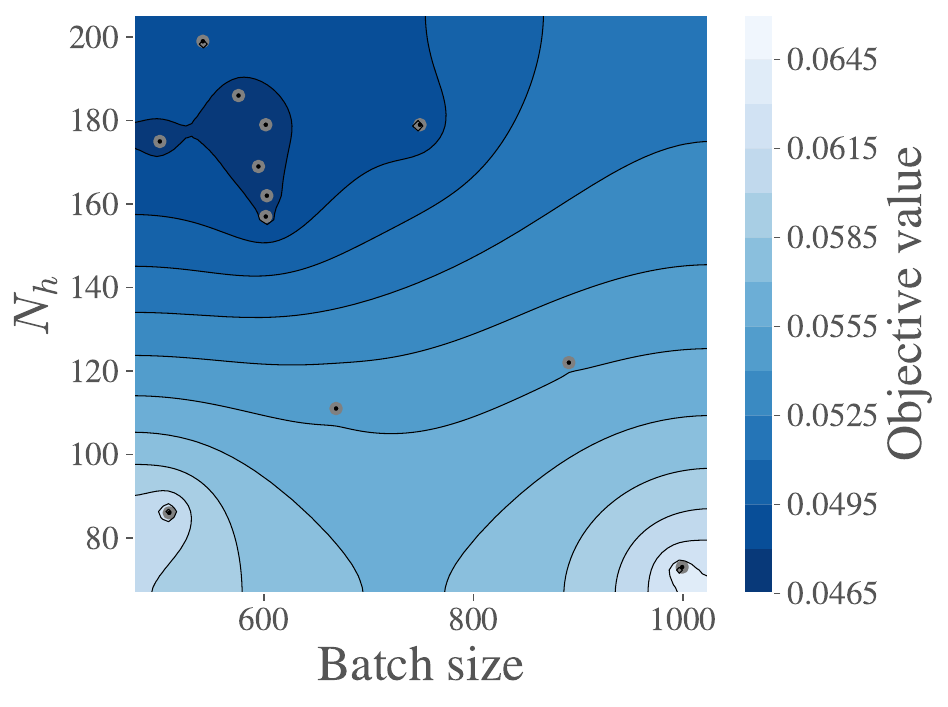}
        \caption{Imaginary component}
        \label{fig:result:pred_u:1-0b}
    \end{subfigure}
    \caption{
    Contours of the objective value (validation error) as a function of hidden size ($N_h$) and batch size for the LSTM training on the (a) real and (b) imaginary components of the latent vector.
    The points represent completed optimization trials.
    }
    \label{fig:result:pred_u:1-0}
\end{figure}

\begin{table}[t]
    \centering
    \caption{Optimized LSTM hyperparameters for the real and imaginary components.}
    \label{tab:result:pred_u:2}
    \begin{tabular}{c|cccccc}
        \hline
        Component & $N_{t,\text{in}}$ & $N_{t,\text{out}}$ & $N_h$ & Batch size & Learning rate & Epochs \\ \hline
        $\mathrm{Re}(\mathbf{z})$ & \multirow{2}{*}{10} & \multirow{2}{*}{1} & 134 & 855 & 0.00084 & \multirow{2}{*}{10000} \\
        $\mathrm{Im}(\mathbf{z})$ & & & 131 & 805 & 0.00057 & \\ \hline
    \end{tabular}
\end{table}

\begin{figure}
    \centering
    \begin{subfigure}{0.49\textwidth}
        \centering
        \includegraphics[width=\textwidth]{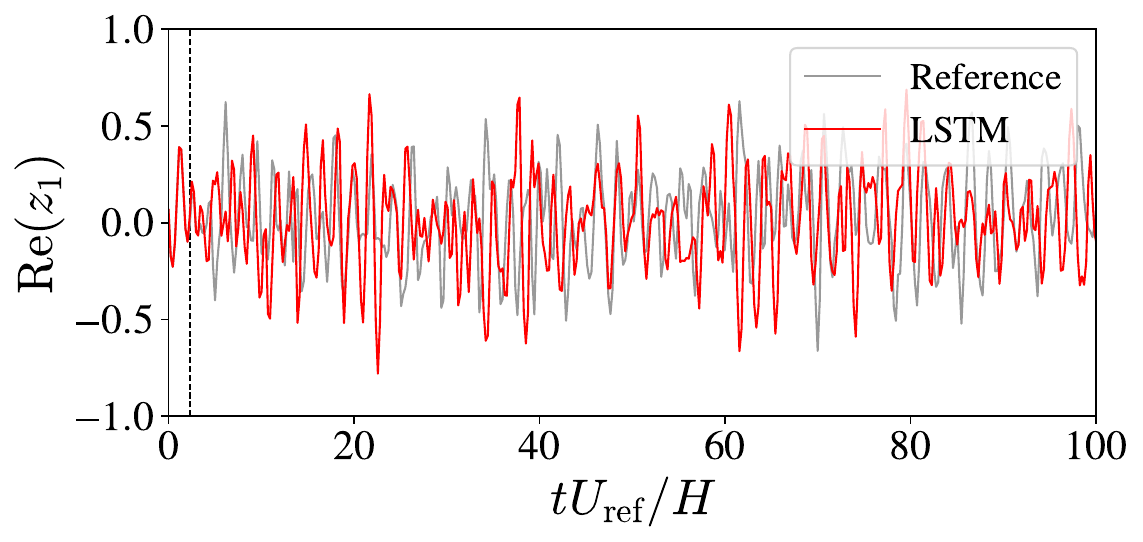}
        \caption{}
        \label{fig:result:pred_u:1a}
    \end{subfigure}
    \begin{subfigure}{0.49\textwidth}
        \centering
        \includegraphics[width=\textwidth]{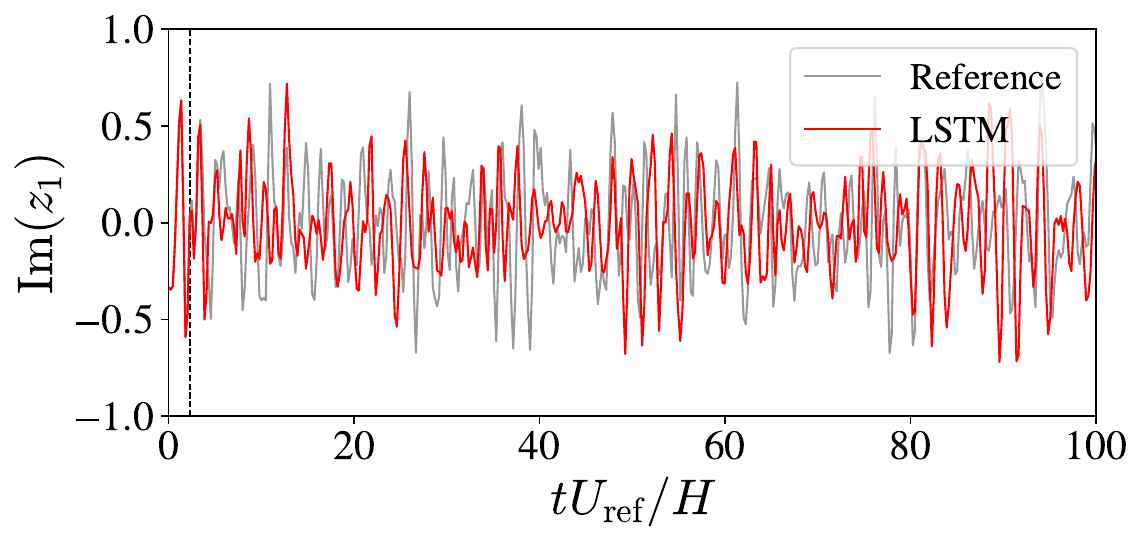}
        \caption{}
        \label{fig:result:pred_u:1b}
    \end{subfigure}
    \caption{
    Comparison of the temporal evolution of the (a) real and (b) imaginary components of the first latent variable $z_{1}(t)$.
    The LSTM forecast (red) is compared against the reference trajectory from the AE (grey).
    The vertical dashed line marks the end of the input sequence ($N_{t,in}$) and the start of the autonomous prediction.
    }
    \label{fig:result:pred_u:1}
\end{figure}

\begin{figure}
    \centering
    \begin{subfigure}{0.45\textwidth}
        \centering
        \includegraphics[width=0.49\textwidth]{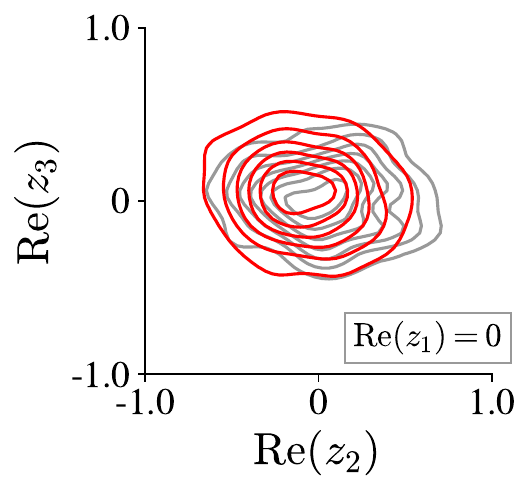}
        \includegraphics[width=0.49\textwidth]{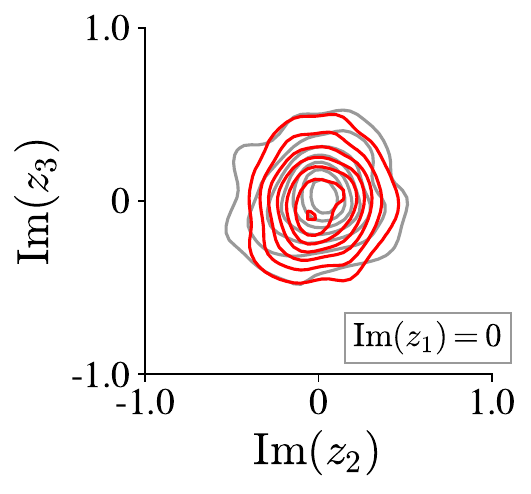}
        \caption{Training data}
        \label{fig:result:pred_u:1-1a}
    \end{subfigure}
    \hspace{3em}
    \begin{subfigure}{0.45\textwidth}
        \centering
        \includegraphics[width=0.49\textwidth]{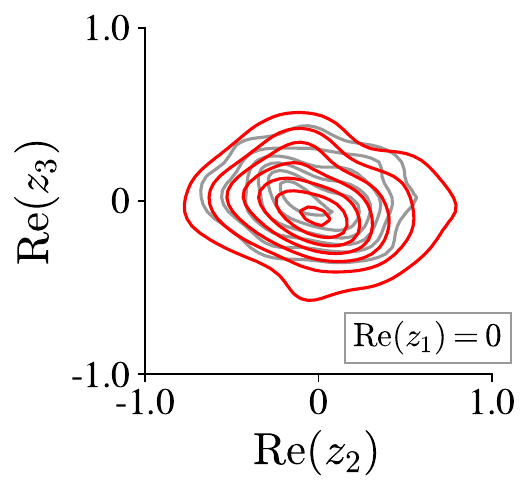}
        \includegraphics[width=0.49\textwidth]{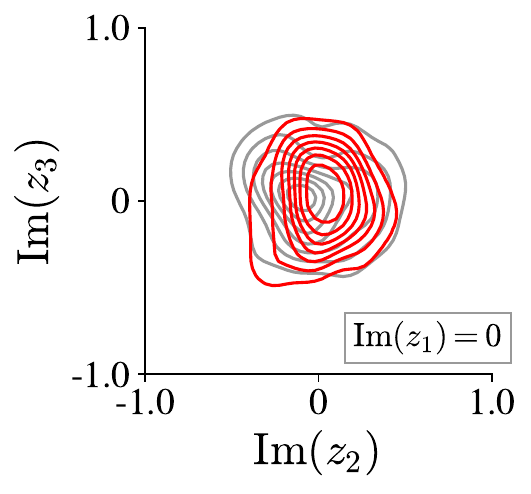}
        \caption{Validation data}
        \label{fig:result:pred_u:1-1b}
    \end{subfigure}
    \caption{
        Poincar\'{e} sections defined by the hyperplane $z_1(t)=0$ with $\diff z_1(t)/\diff t > 0$.
        The contours represent the joint Probability Density Function (PDF) of the intersection points in the $z_2$-$z_3$ phase plane.
        (a) Training set, and (b) Validation set. Gray contours: Reference; Red contours: LSTM prediction.
    }
    \label{fig:result:pred_u:1-1}
\end{figure}

\begin{figure}
    \centering
    \begin{subfigure}{0.49\textwidth}
        \centering
        \includegraphics[width=\textwidth]{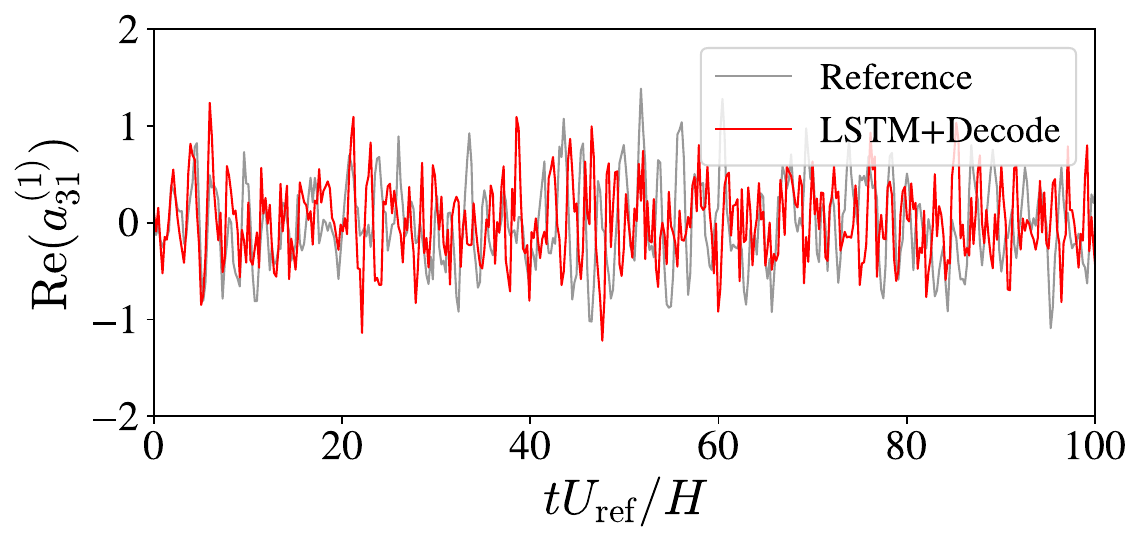}
        \caption{Training window}
        \label{fig:result:pred_u:2a}
    \end{subfigure}
    \begin{subfigure}{0.49\textwidth}
        \centering
        \includegraphics[width=\textwidth]{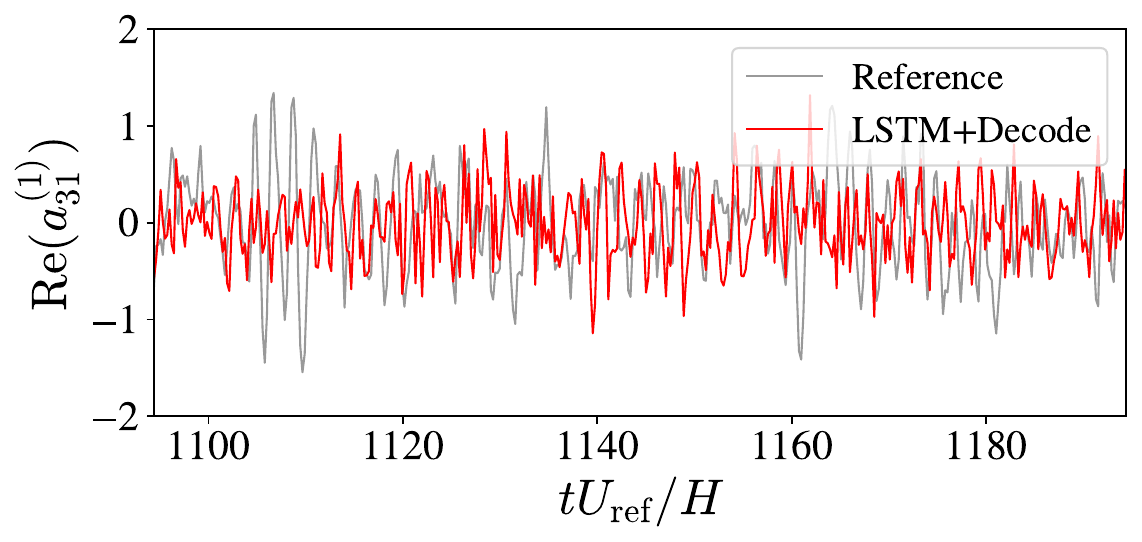}
        \caption{Test window}
        \label{fig:result:pred_u:2b}
    \end{subfigure}
    \\
    \begin{subfigure}{\textwidth}
        \centering
        \includegraphics[width=0.8\textwidth]{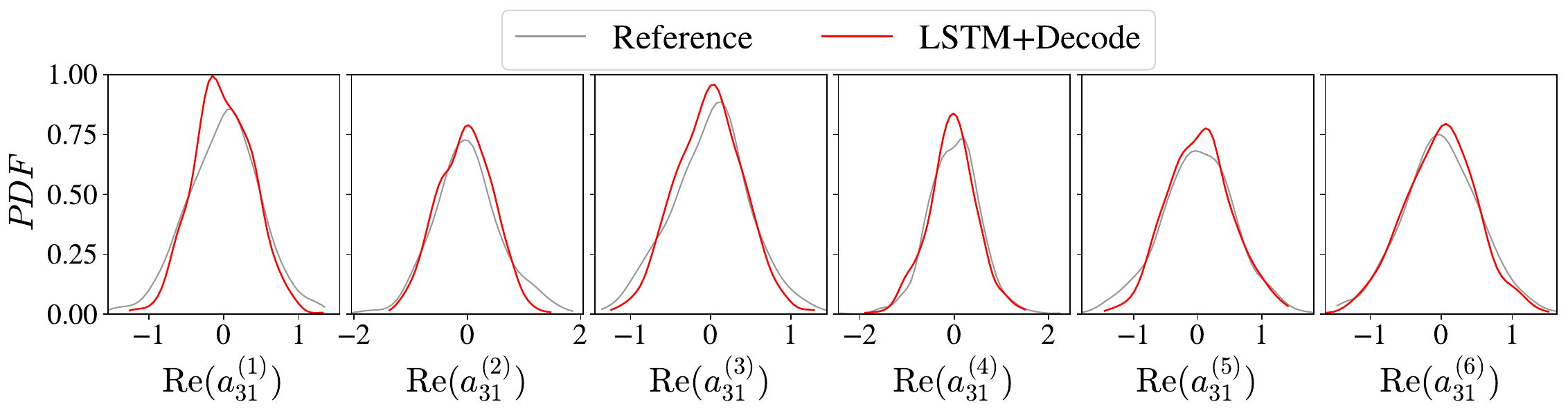}
        \caption{PDF statistics (test window)}
        \label{fig:result:pred_u:2c}
    \end{subfigure}
    \caption{
    (a, b) Temporal evolution of the predicted (red) and reference (grey) real component of the SPOD coefficient $a_{31}^{(1)}(t)$ in the (a) training and (b) testing windows.
    (c) Probability density functions (PDFs) of the predicted coefficients for selected dominant modes in the test window.
    }
    \label{fig:result:pred_u:2}
\end{figure}

\begin{figure}
    \centering
    \begin{subfigure}{0.49\textwidth}
        \centering
        \includegraphics[width=\textwidth]{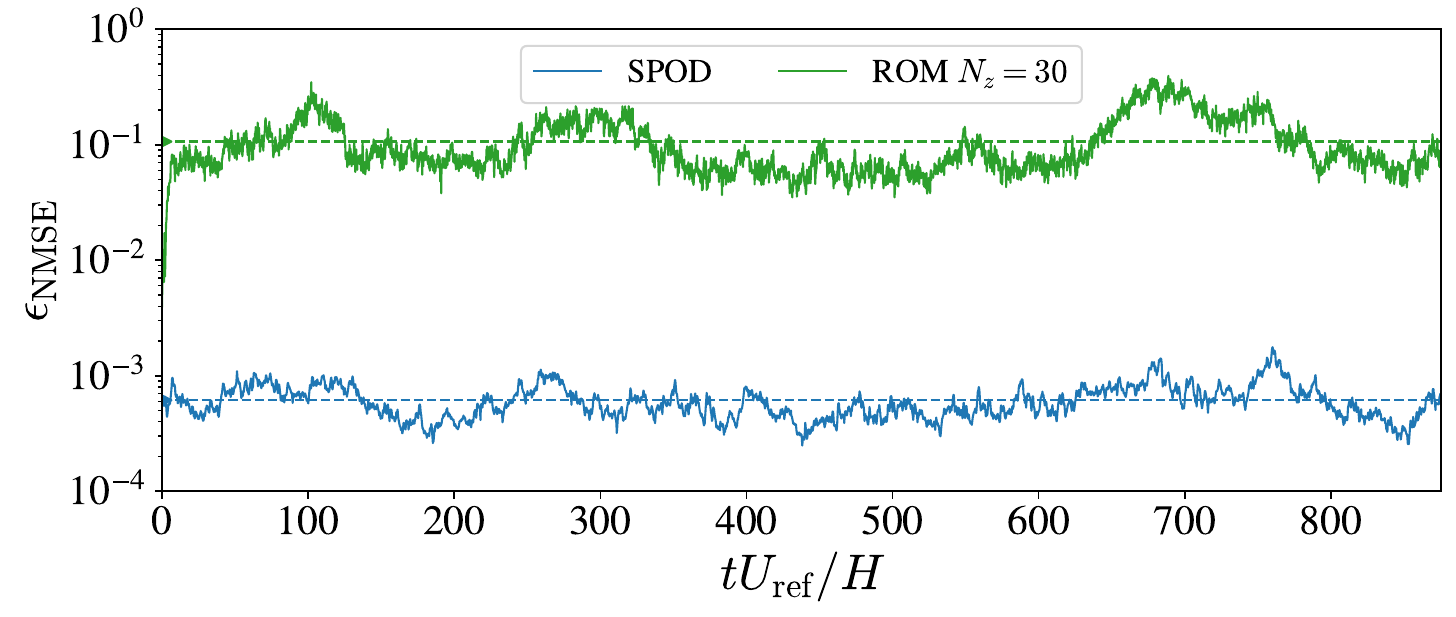}
        \caption{Training window}
        \label{fig:result:pred_u:3a}
    \end{subfigure}
    \begin{subfigure}{0.49\textwidth}
        \centering
        \includegraphics[width=\textwidth]{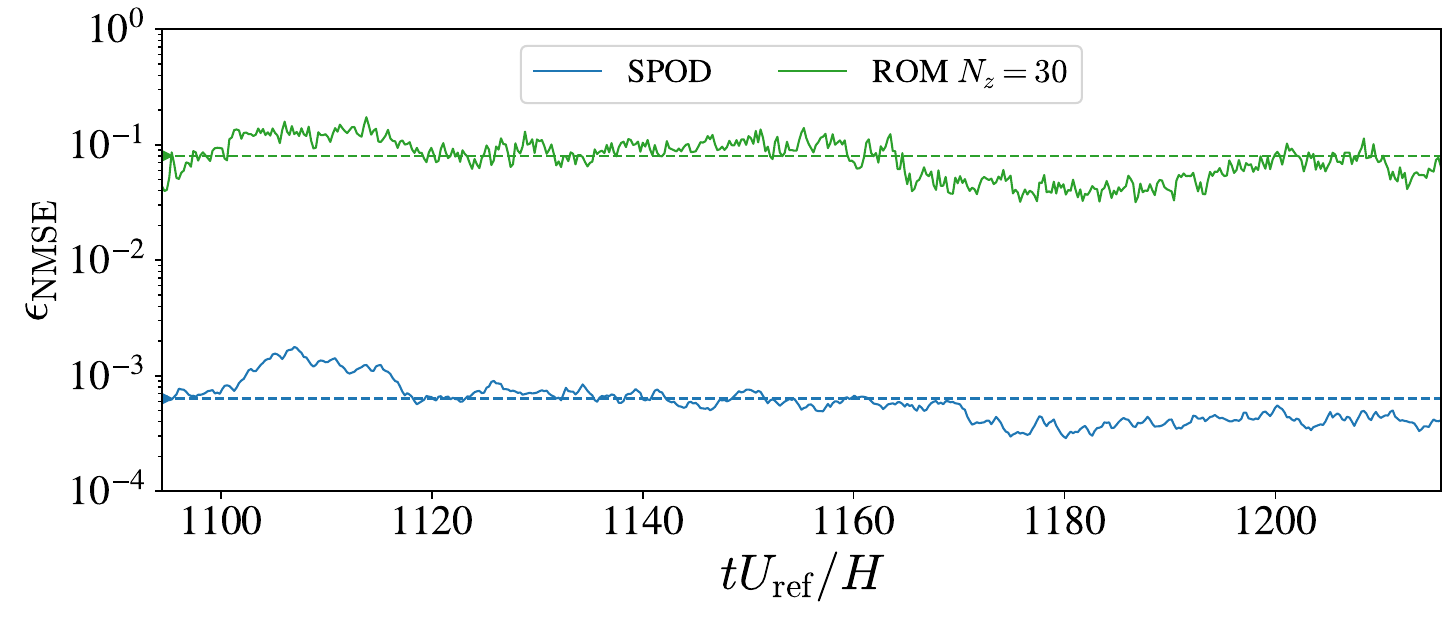}
        \caption{Test window}
        \label{fig:result:pred_u:3b}
    \end{subfigure}
    \caption{
    Time evolution of the normalized mean-squared error (NMSE) of the velocity magnitude field.
    The error is computed between the LSTM-ROM reconstruction and the reduced SPOD reference.
    Dashed lines indicate the time-averaged error.
    }
    \label{fig:result:pred_u:3}
\end{figure}

To perform temporal forecasting, a surrogate model is required to propagate the compressed latent state $\mathbf{z}(t) \in \mathbb{C}^{N_z}$ forward in time.
This propagated state is subsequently decoded by the AE and projected back onto the SPOD basis to reconstruct the full velocity field.
The long short-term memory (LSTM) network, detailed in \sref{sec:method:lstm}, is selected for this task.

The model is implemented using the \texttt{Darts} library \citep{herzenDartsUserFriendlyModern2022}.
Separate networks are trained for the real and imaginary components of the latent vector.
The training dataset is split into $80\%$ for gradient descent and $20\%$ for validation.
The input sequence length is fixed at $N_{t,\text{in}} = 10$, with a forecast horizon of $N_{t,\text{out}} = 1$.
Hyperparameter optimization is conducted using the \texttt{Optuna} framework \citep{akibaOptunaNextgenerationHyperparameter2019}.
A search space is defined for the hidden layer size ($N_h$), batch size, and learning rate, as summarized in \tref{tab:result:pred_u:1}.
Twenty trials of 100 epochs each are executed.
Figure \ref{fig:result:pred_u:1-0} illustrates the sensitivity of the validation error to $N_h$ and batch size.
The optimal hyperparameters, corresponding to validation errors of $0.060$ (real) and $0.047$ (imaginary), are listed in \tref{tab:result:pred_u:2}.

The trained LSTM is deployed in a recursive configuration to generate autonomous forecasts.
The ability of the model to capture the turbulent dynamics is demonstrated in \fref{fig:result:pred_u:1}, which compares the predicted trajectory of the first latent variable $z_1(t)$ against the reference trajectory.
As expected for a chaotic system, the point-wise trajectory diverges after several characteristic time units.
However, the prediction remains bounded within the correct dynamical range.
To rigorously validate the learned physics, Poincar\'{e} maps were computed \citep{solera-ricoVVariationalAutoencodersTransformers2024}.
Intersections of the trajectory with the hyperplane $z_1(t)=0$ (where $\diff z_1/\diff t > 0$) were recorded, and the joint probability density function (PDF) of the intersections in the $z_2$-$z_3$ plane was estimated.
As shown in \fref{fig:result:pred_u:1-1}, the LSTM accurately reproduces the topology of the attractor in the phase space for both the training and validation sets.
This confirms that the model has learned the underlying invariant statistics of the flow, rather than simply memorizing a specific trajectory.

The forecasted latent vectors are decoded to reconstruct the SPOD coefficients.
Figures \ref{fig:result:pred_u:2a} and \ref{fig:result:pred_u:2b} display the trajectories of a representative coefficient in the training and testing windows, respectively.
In the test window, the prediction is initialized using only the last $N_{t,\text{in}}$ steps of the training data.
Despite the divergence in phase, the statistical moments are well-preserved, as evidenced by the matching PDFs in \fref{fig:result:pred_u:2c}.
Finally, the full velocity field is reconstructed.
The time evolution of the reconstruction error is shown in \fref{fig:result:pred_u:3}.
While the error grows as the forecast horizon extends, it saturates at a bounded level in the test window, confirming the long-term stability of the ROM.

% ---------------------------------------------------------------------------
\subsection{Prediction of pollutant field}
% ---------------------------------------------------------------------------
\label{sec:result:pred_c}

\begin{table}[t]
    \centering
    \caption{CNN hyperparameters for the velocity-to-concentration mapping.}
    \label{tab:result:pred_c:1}
    \begin{tabular}{ccccccc}
        \hline
        Channels & Filter size & Kernel size & Padding & Batch size & Learning rate & Epochs \\ \hline
        [2, 32, 64, 128, 64, 32, 1] & $96 \times 96$ & 5 & 2 & 16 & 0.001 & 1000 \\ \hline
    \end{tabular}
\end{table}

\begin{figure}
    \centering
    \includegraphics[width=.8\textwidth]{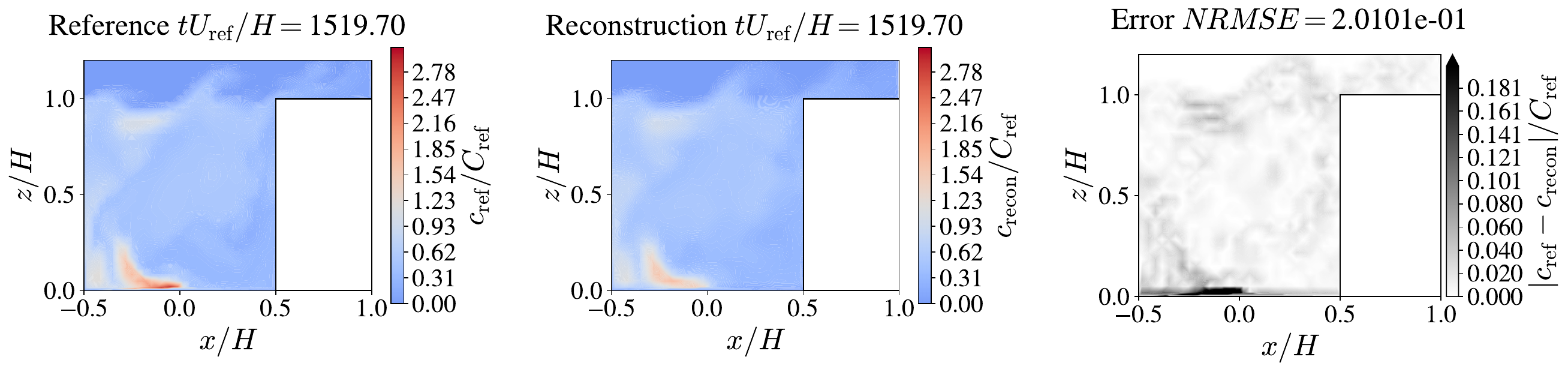}
    \caption{
    Instantaneous reconstructed concentration field obtained from the CNN in the validation window.
    }
    \label{fig:result:pred_c:1}
\end{figure}

\begin{figure}
     \centering
    \includegraphics[width=.55\textwidth]{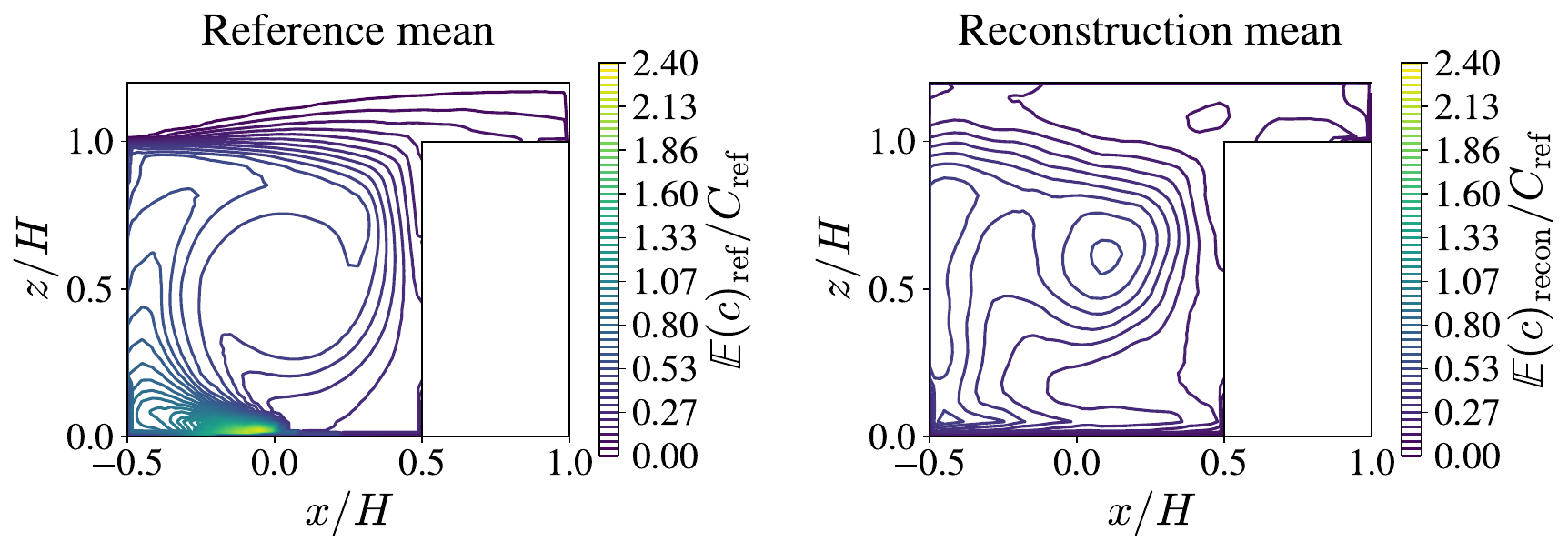}
    \caption{
    Contours of the mean concentration field computed over the test window.
    }
    \label{fig:result:pred_c:2}
\end{figure}

\begin{figure}
     \centering
    \begin{subfigure}{0.4\textwidth}
        \centering
        \includegraphics[width=\textwidth]{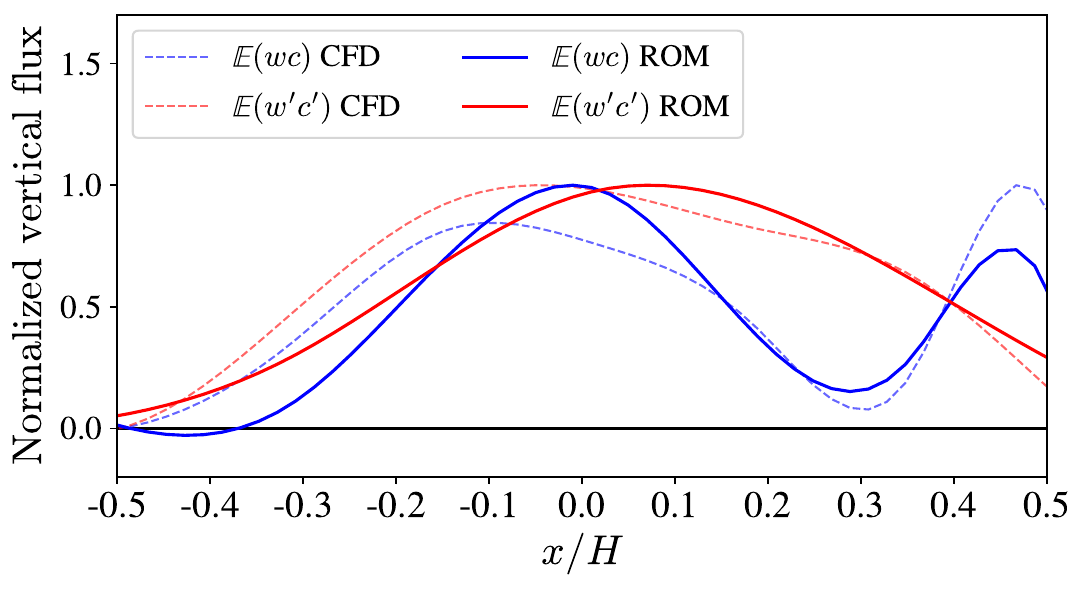}
        \caption{}
        \label{fig:result:pred_c:3a}
    \end{subfigure}
    \begin{subfigure}{0.59\textwidth}
        \centering
        \includegraphics[width=\textwidth]{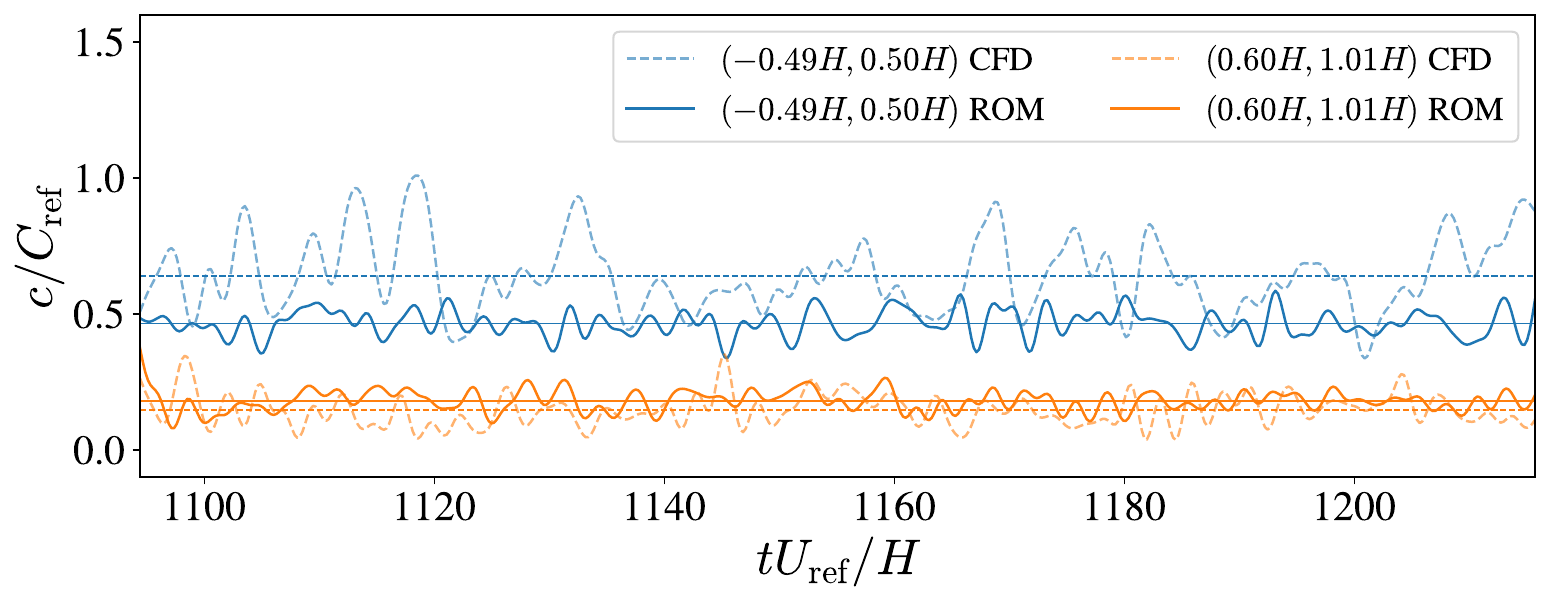}
        \caption{}
        \label{fig:result:pred_c:3b}
    \end{subfigure}
    \caption{
    (a) Comparison of the mean total vertical mass flux of the pollutant across the canopy interface ($z/H=1$).
    (b) Time evolution of pollutant concentration at two probe locations: Outside the canyon ($x=0.60H, z=1.01H$) and inside the canyon on the leeward face ($x=-0.49H, z=0.50H$).
    }
    \label{fig:result:pred_c:3}
\end{figure}

To reconstruct the pollutant concentration field, a nonlinear mapping from the velocity field is learned using the convolutional neural network (CNN) architecture described in \sref{sec:method:cnn}.
The network is trained in a supervised manner using the high-fidelity LES fields.
The specific architectural parameters and hyperparameters used for the training are summarized in \tref{tab:result:pred_c:1}.
A preliminary sensitivity analysis confirmed that the model performance is robust to minor variations in these parameters; consequently, the configuration presented here is adopted for the final results.

The fidelity of the instantaneous reconstruction is visualized in \fref{fig:result:pred_c:1} in the validation window.
While the macro-scale topology of the plume is well-captured, the model exhibits a smoothing effect on the fine-scale structures, particularly near the point source at the canyon floor.
This behavior is characteristic of regression models trained with a mean-squared error (MSE) loss function, which tends to filter out high-frequency spatial gradients.
Nevertheless, the overall agreement with the reference field is sufficient for applications where the robust estimation of macro-scale dispersion features is the primary objective.

Three key metrics were analyzed to assess the ROM's capability in the testing window (using predicted velocity fields).
First, the time-averaged concentration field is computed and compared with the CFD reference in \fref{fig:result:pred_c:2}.
Despite the smoothing of the shear layer and the source region, the primary transport mechanism—the downward advection of pollutants along the windward wall driven by the cavity recirculation—is accurately reproduced.
This agreement indicates that the error accumulation in the velocity forecast (discussed in \sref{sec:result:pred_u}) does not drastically degrade the scalar field prediction.

Second, the vertical exchange of pollutants, quantified by calculating the mean total mass flux across the roof level ($z/H=1$), is shown in \fref{fig:result:pred_c:3a}.
The underprediction observed in the upstream region ($x/H < 0$) is attributed to the CNN's difficulty in resolving the sharp gradients of the shear layer.
However, the qualitative variation of the flux across the canyon width is captured, confirming that the ROM preserves the physics of the ventilation mechanism.

Finally, the temporal fidelity is evaluated by monitoring concentration histories at two distinct probes (\fref{fig:result:pred_c:3b}).
At the outer probe (located above the shear layer), the ROM successfully captures the fluctuation dynamics.
At the inner probe (located in the recirculation zone), the concentration levels are underpredicted by approximately $16\%$, consistent with the smoothing of the source intensity.
Overall, these results confirm that the offline ROM serves as a computationally efficient surrogate for estimating pollutant transport, providing actionable insights at a fraction of the cost of full CFD simulations.

%% file: section-conclusion.tex
% ---------------------------------------------------------------------------
\section{Conclusion}
% ---------------------------------------------------------------------------
\label{sec:conc}

A data-driven ROM framework based on SPOD is proposed for modeling the urban street canyon flow simulated by LES, with a focus on the pollutant dispersion.
A novel dimensionality reduction technique based on mode energy and spatial similarity has been introduced in the SPOD modal space.
The time-domain coefficients corresponding to the preserved modes has been used to train a neural-network framework for dynamical prediction.
This framework consisted of an AE network for data compression, a LSTM network for evolution of the latent space in time, and a CNN for transformation of the velocity field to concentration field.
The ROM has been applied for offline inference where flow field data, unseen during the training phase, has been used for prediction of velocity and concentration fields.

The SPOD dimensionality reduction offered a $97\%$ reduction over the dimension of the full SPOD space while preserving the spectral and spatial flow features.
The AE compression provided a latent space with a dimension lower than the space of preserved SPOD coefficients by a factor of $66$, while reproducing the temporal behavior.
The LSTM prediction of the SPOD coefficients was able to emulate the reference dynamics over a time horizon unseen during the training.
The deviation of the predicted velocity fields using the forecasted temporal coefficients from the CFD data remained uniform and bounded through the prediction horizon.
The velocity fields mapped to concentration field using the trained CNN was able to replicate the large scale principal recirculation region in the canyon, the characteristic mass flux profile, and pollutant levels at fixed locations in the flow field.
In conclusion, the novel data-driven ROM offers an efficient framework for accurate and stable velocity and pollutant predictions, offering a potential tool to facilitate urban planning.

It is acknowledged that the success of the proposed framework relies on the ability of the SPOD modes to span the possible range of parameters encountered in the application.
In order to ensure robustness in parametric variability, more configurations must be incorporated in the training database and the model performance must be verified.
Also, in terms of computational cost, training the neural networks is the most demanding.
However, it is only required to be performed once, and a possibility to incorporate future observations by either using data assimilation to update the model output, or using transfer learning to retrain the neural networks at a reduced computational cost, can be explored.